\documentclass[journal]{IEEEtran}

\usepackage{cite}

\usepackage{color,xcolor}

\usepackage{graphicx}
\usepackage{subfigure}

\usepackage{bm}		
\usepackage{amsmath}
\usepackage{amsfonts}	
\usepackage{enumerate}

\usepackage{makecell}
\usepackage{threeparttable}
\usepackage{footnote}

\usepackage{algpseudocode}
\usepackage{algorithm}

\begin{document}
	\title{An Adaptive Neuro-Fuzzy System with Integrated Feature Selection and Rule Extraction for High-Dimensional Classification Problems}
	
	\author{
		Guangdong Xue, Qin Chang, Jian Wang, \IEEEmembership{Senior Member, IEEE}, Kai Zhang, and Nikhil R. Pal, \IEEEmembership{Fellow, IEEE}
		
		\thanks{This work was supported in part by the National Key Research and Development Program of China under Grant 2018AAA0100100; in part by the National Natural Science Foundation of China under Grant 62173345; in part by the Major Scientific and Technological Projects of China National Petroleum Corporation (CNPC) under Grant ZD2019-183-008; and in part by the Fundamental Research Funds for the Central Universities under Grant 20CX05002A and Grant 20CX05012A.
			
		G. Xue is with the College of Control Science and Engineering, China University of Petroleum (East China), Qingdao 266580, China (e-mail: g.d.xue@foxmail.com).
		
		Q. Chang and J. Wang are with the College of Science, China University of Petroleum (East China), Qingdao 266580, China (e-mail: chq\_@163.com; wangjiannl@upc.edu.cn).
			
		K. Zhang is with the College of Petroleum Engineering, China University of Petroleum (East China), Qingdao 266580, China (e-mail: zhangkai@upc.edu.cn).
			
		N. R. Pal is with the Electronics and Communication Sciences Unit, Indian Statistical Institute, Calcutta, 700108, India (e-mail: nrpal59@gmail.com).}
	}
	
	\markboth{Manuscript for IEEE Transactions on Fuzzy Systems}
	{Shell \MakeLowercase{\textit{et al.}}: Bare Demo of IEEEtran.cls for IEEE Journals}
	
	\maketitle
	
	\begin{abstract}
		A major limitation of fuzzy or neuro-fuzzy systems is their failure to deal with high-dimensional datasets. This happens primarily due to the use of T-norm, particularly, product or minimum (or a softer version of it). Thus, there are hardly any work dealing with datasets with dimensions more than hundred or so. Here, we propose a neuro-fuzzy framework that can handle datasets with dimensions even more than 7000! In this context, we propose an adaptive softmin (Ada-softmin) which effectively overcomes the drawbacks of ``numeric underflow" and ``fake minimum" that arise for existing fuzzy systems while dealing with high-dimensional problems. We call it an Adaptive Takagi-Sugeno-Kang (AdaTSK) fuzzy system. We then equip the AdaTSK system to perform feature selection and rule extraction in an integrated manner. In this context, a novel gate function is introduced and embedded only in the consequent parts, which can determine the useful features and rules, in two successive phases of learning. Unlike conventional fuzzy rule bases, we design an enhanced fuzzy rule base (En-FRB), which maintains adequate rules but does not grow the number of rules exponentially with dimension that typically happens for fuzzy neural networks. The integrated Feature Selection and Rule Extraction AdaTSK (FSRE-AdaTSK) system consists of three sequential phases: (i) feature selection, (ii) rule extraction, and (iii) fine tuning. The effectiveness of the FSRE-AdaTSK is demonstrated on 19 datasets of which five are in more than 2000 dimension including two with dimension greater than 7000. This may be the {\it first time} fuzzy systems are realized for classification involving more than 7000 input features.
	\end{abstract}
	
	
	\section{Introduction}\label{introduction}
	\IEEEPARstart{F}{uzzy} systems have been successfully applied in many areas, such as control engineering \cite{1975MamdaniAn,1985TakagiFuzzy,1986SugenoFuzzy,2001PalIntegrated} and pattern recognition \cite{2004PalA,2012PalAnIntegrated,2018PalFeature,2019GaoConjugate,2020WuOptimize,2020CuiOptimize}. Takagi-Sugeno-Kang (TSK) fuzzy systems \cite{1985TakagiFuzzy,1986SugenoFuzzy}, one of two classical fuzzy systems (the other one is Mamdani-Assilian (MA) fuzzy system \cite{1975MamdaniAn}), have powerful ability to model nonlinear problems. To overcome the difficulties in identifying the fuzzy rules, various neural networks have been proposed based on their promising learning performance \cite{1992HorikawaOn,2001PalIntegrated,2004PalA,2018PalFeature,2020WuOptimize,2020CuiOptimize}. As a consequence, back-propagation algorithm \cite{1986RumelhartLearning} has become a popular scheme to optimize TSK fuzzy systems.
	
	The computation of the firing strength of fuzzy rules is a necessary and important operation in fuzzy reasoning. The firing strength can be computed using any T-norm \cite{1995KlirFuzzy}. In \cite{1989MizumotoPictorial}, two kinds of T-norms have been used which include minimum \cite{1985TakagiFuzzy} and product \cite{2020CuiOptimize,2020WuOptimize,2018PalFeature,2021GuFast}. The product T-norm is differentiable and frequently used for solving low-dimensional problems. Since minimum is not differentiable, it is rarely considered when using gradient based algorithm for learning parameters of fuzzy systems. To remedy this drawback, different softer versions of minimum called softmin have been employed \cite{2004PalA,2008PalSimultaneous,2019GaoConjugate} although strictly speaking they are not T-norms. The softmin is differentiable and can be a good approximator to minimum. In \cite{2008PalSimultaneous}, the authors adopted the product and softmin to compute the firing strength, separately. The simulation results revealed that these two T-norm based systems performed similarly on four regression datasets. For high-dimensional problem, one can easily deduce that the firing strength computed by the product T-norm would reduce almost to zero even when each antecedent component has a high membership value. Sometimes, this phenomenon is called  ``numeric underflow'' \cite{2020CuiOptimize}, which means that the firing strength is too small to be correctly represented and used by the machine. The softmin is also prone to suffering from the same obstacle, ``numeric underflow''. It also has another problem, which may be called as ``fake minimum'' when the parameter of the softmin is not properly set. Thus, it is very important to investigate how we can design a novel adaptive version of softmin which can avoid these shortcomings to deal with high-dimensional problems.
	
	For solving high-dimensional classification problems, feature selection is of significant necessity because when the input dimension is high, it is very likely that the input feature set contains redundant and derogatory features, which are expected to negatively affect the final performance \cite{2021WangFeature}. In addition to improving the system performance, feature selection can also simplify the complexity, lower the computational cost, and enhance the interpretability of the system. Generally, feature selection methods are  classified into three groups: filter methods \cite{2017SolorioANew}, wrapper methods \cite{1997KohaviWrappers,2009MaldonadoA,2010KabirA}, and embedded methods \cite{2012PalAnIntegrated,2018PalFeature}. A filter method selects features according to an evaluation criteria such as mutual information or correlation which is independent of the classifier or predictor, that will finally use these features. Although, filter methods are efficient in computation, they may result in unsatisfactory results because they ignore the feedback of the model \cite{2011HsuHybrid} as well as the interaction between features. A wrapper method considers different feature subsets and then evaluates the goodness of them using classification or prediction accuracy. This results in good performance but costs more computational resource \cite{2019ChenAn}. Moreover, to get the best subset, we need evaluate all possible subsets of features, which is not feasible for high-dimensional datasets. An embedded method  \cite{2012PalAnIntegrated} has gradually been an attractive research topic. It simultaneously performs the feature selection and evaluation of the model and consequently, it promotes the effectiveness and reduces the computational burden. Additionally, some hybrid methods \cite{2015HuHybrid,2018LiuA} which utilize two of these three feature selection approaches together have also been investigated. In \cite{2015HuHybrid}, the Partial Mutual Information based filter approach was coupled with a firefly algorithm based wrapper method to deal with short-term load forecasting problem. And in \cite{2018LiuA}, one hybrid wrapper-embedded feature selection method was proposed combining genetic algorithm and the embedded regularization approach.
	
	Basically, there are four kinds of features \cite{2008PalSelecting}: essential features, redundant features, derogatory or bad features, and indifferent features. Essential features are of great importance to reach competitive performance, while derogatory or bad features have adverse effects. Indifferent features have neither positive nor negative influence on the tasks, and as for redundant features, they are useful but not all of them are needed. Feature selection method should select essential features, discard bad and indifferent features, and control the use of redundant features \cite{2015PalFeature}. In \cite{2018PalFeature} and \cite{2021WangFeature}, features are selected using a fuzzy rule based framework and neural networks, respectively, and both of these  works considered controlling the level of redundancy among selected features. In \cite{2012PalAnIntegrated} and \cite{2020ZhangFeature}, the authors did not take into consideration redundant features and got comparable results as well.
	
	In \cite{2017SolorioANew}, an unsupervised spectral feature selection method was designed by combining a kernel with a specific spectrum based on a feature evaluation metric. Compared with other filter methods, the experimental results demonstrated promising performance. Kabir \textit{et al.} \cite{2010KabirA} integrated a wrapper approach into neural networks which automatically determined the architecture of the network model during the feature selection process. An embedded feature selection method was proposed in \cite{2020ZhangFeature,2021WangFeature}, which added the group lasso regularization to the loss function of neural networks to select useful features simultaneously with the training of the network. In \cite{2001PalIntegrated,2004PalA,2008PalSimultaneous} and \cite{2008PalSelecting}, feature modulators were designed and used to modify the membership values in antecedent parts of fuzzy rules for feature selection. Note that the feature modulator was introduced as gate function in \cite{2008PalSimultaneous} which modified both antecedent and consequent parts, and finally picked up the required features. Actually, there are three commonly used gate functions to measure the magnitude of feature's importance, $1-e^{-\lambda^2}$ (\cite{2001PalIntegrated,2004PalA}), $\frac{1}{1+e^{-\lambda}}$ and $e^{-\lambda^2}$ (\cite{2012PalAnIntegrated,2015PalFeature,2018PalFeature}), where $\lambda$ is the tuneable gate parameter. In the existing works, the gate values (the values of gate function) are generally initialized around zeros, and the corresponding derivatives are close to zeros as well. This leads to more training iterations to open the gates during feature selection. Thus, designing of a suitable gate function to overcome this drawback is another interesting task.

	How to generate or extract desirable fuzzy rules is an important part in constructing a fuzzy system. A variety of approaches have been introduced, such as heuristic approaches \cite{1995AbeA,1992IshibuchiDistributed}, genetic algorithms \cite{1995IshibuchiSelecting,2004IshibuchiFuzzy,2006IshibuchiFuzzy}, and neuro-fuzzy techniques \cite{2004PalA,2008PalSimultaneous}. In \cite{1995AbeA}, an efficient method was proposed to extract fuzzy rules directly from input data through activation and inhibition hyperboxes. Ishibuchi et al. \cite{1992IshibuchiDistributed} used a distributed representation of fuzzy rules for classification of patterns. This approach used multiple fuzzy partitions simultaneously to generate rules for classification. The advantage of this method is that it does not need any iterative computations and complex procedures. In \cite{2004IshibuchiFuzzy}, Ishibuchi and Yamamoto proposed a two-phase method to select a small number of simple rules. In phase I, the candidate rules were generated using two metrics (\textit{confidence} and \textit{support}) and in phase II, the appropriate fuzzy rules were selected by a multi-objective genetic local search algorithm. In a neuro-fuzzy framework \cite{2004PalA}, all possible fuzzy rules were initially employed but after the training only an adequate set of rules involving a set of selected features was retained.
	
	Depending on the method of extraction, fuzzy rule bases can be divided into two categories, compactly combined fuzzy rule base (CoCo-FRB) and fully combined fuzzy rule base (FuCo-FRB). After the feature space is clustered or partitioned, CoCo-FRB converts each cluster into one fuzzy rule, in other words, the number of rules is equal to the number of clusters. While FuCo-FRB considers all possible rules. Here on the domain of each feature,  a number of linguistic values are defined and all possible valid combinations that can define rules are considered. Hence, the number of rules in FuCo-FRB exponentially increases with the dimension. It is still a challenging task to address the issue of exponential increase in the number of rules when one employs FuCo-FRB especially for high-dimensional dataset. In \cite{2020WuOptimize}, a promising scheme was proposed to effectively address this problem. It first used principle component analysis (PCA) to reduce the original dimension to $5$ and then the rule base was initialized to FuCo-FRB. Inspired by DropOut \cite{2014SrivastavaDropout} and DropConnect \cite{2013WanRegularization}, the so-called DropRule strategy was adopted in each iteration which randomly discarded the rules based on the drop rate. The simulation results demonstrate the superiority over its counterparts. Unfortunately, this scheme inevitably affect the interpretability of the constructed system.
	
	\begin{table*}[t]
		\caption{Main notations used in this paper}
		\begin{center}
			\begin{tabular}{c | l}
				\hline
				Notation & \makecell[c]{Description}\\
				\hline
				$D$ & The number of features\\
				$S$ & The number of fuzzy sets for each feature\\
				$C$ & The number of classes in the dataset\\
				$R$ & The number of rules in the TSK fuzzy system\\
				$\mathbf{x}=(x_1,x_2,\cdots,x_D)^T$ & $D$-dimensional feature vector\\
				$m_{r,d}$, $\sigma_{r,d}$ & Center and spread of membership function for the $d_{th}$ feature in the $r_{th}$ rule\\
				$p_{r,d}^{c}$ & Consequent parameter of the $r_{th}$ rule associated with the $d_{th}$ feature for the $c_{th}$ class\\
				$A_{r,d}$ & The fuzzy set associated with the $d_{th}$ feature in the $r_{th}$ rule\\
				$\mu_{r,d}(\mathbf {x})$ & The membership value of the $d_{th}$ feature of $\mathbf{x}$ in the $r_{th}$ rule\\
				$f_r(\mathbf{x})$, $\bar{f_r}(\mathbf{x})$ & The firing strength and normalized firing strength of the $r_{th}$ rule for $\mathbf{x}$\\
				$y_r^c(\mathbf{x})$ & The output of the $r_{th}$ rule of the $c_{th}$ class for $\mathbf{x}$\\
				$y^c(\mathbf{x})$ & The output of the TSK fuzzy system of the $c_{th}$ class for $\mathbf{x}$\\
				$y_c(\mathbf{x})$ & The $c_{th}$ component of true label of $\mathbf{x}$ coded by one-hot coding\\
				$E$ & System error\\
				$N$ &  The number of training samples\\
				$\hat{q}$ & The parameter of Ada-softmin\\
				$\eta$ & The learning rate\\
				$I$ & The index matrix to calculate firing strength\\
				$\lambda$, $\theta$ & The parameter of gate function for feature selection and rule extraction\\
				$\tau$ & The threshold of feature selection and rule extraction\\
				$\zeta$ & The coefficient to compute threshold\\
				\hline
			\end{tabular}
		\end{center}
		\label{notation}
	\end{table*}

	In this paper, we focus on TSK neuro-fuzzy systems to deal with high-dimensional classification problems. We introduce an  adjustable-parameter-based softmin to compute the firing strength. Additionally, we propose a gate function to realize feature selection and rule extraction. As a result, we obtain a comprehensive feature selection and rule extraction-based system, that we call FSRE-AdaTSK. FSRE-AdaTSK consists of three sequential phases: (i) feature selection, (ii) rule extraction, and (iii) fine tuning. The main contributions in this work are summarized as follows:
	
	\begin{itemize}
		\item An adaptive softmin called as Ada-softmin is proposed whose index parameter is adjusted on the basis of current membership values. For every rule and every data point, depending on the present membership values of the antecedent clauses, the parameter of the Ada-softmin changes. It is adopted to compute the firing strength in the TSK fuzzy system which effectively avoids the two typical problems: ``numeric underflow'' and ``fake minimum''. An adaptive TSK (AdaTSK) neuro-fuzzy system is then derived which possesses the ability to deal with high-dimensional data.
		\item Inspired by the existing gate functions \cite{2001PalIntegrated,2004PalA,2012PalAnIntegrated,2015PalFeature,2018PalFeature}, a novel gate function is designed. When the gate values are initialized around zeros, the derivatives of the proposed gate function are much greater than those of the existing ones. This enables the proposed TSK neuro-fuzzy system to efficiently accomplish the feature selection task. 
		\item Based on CoCo-FRB, we obtain an enhanced fuzzy rule base (En-FRB). It collects more fuzzy rules than CoCo-FRB and avoids the exponential increase in the number of rules caused by FuCo-FRB. Finally, for the constructed En-FRB, we propose an efficient embedded rule extraction method using the above presented gate function.
		\item This results in a neuro-fuzzy classifier that can deal with even datasets having dimension greater than 7000.
		\item The proposed framework can be easily adapted to fuzzy rule based systems.
	\end{itemize}
	
	The rest of this article is organized as follows. CoCo-FRB based TSK fuzzzy system is reviewed and the AdaTSK method is introduced in the next section. Section \ref{FSRE_AdaTSK_method} elaborates the details of the proposed FSRE-AdaTSK system. Section \ref{experiments_results} demonstrates the effectiveness of the approach proposed in this paper. In Section \ref{conclusion}, our conclusions are provided and possible future research directions are also discussed.

	\section{TSK Fuzzy System with Adaptive Softmin (AdaTSK)}
	In this section, we will first review the CoCo-FRB based TSK fuzzy system for classification problems and then introduce an adaptive softmin to construct the so-called AdaTSK system. For convenience, the notations used in this paper are listed in Table \ref{notation}.

	\begin{figure*}[t]
		\centering
		\includegraphics[scale=0.5]{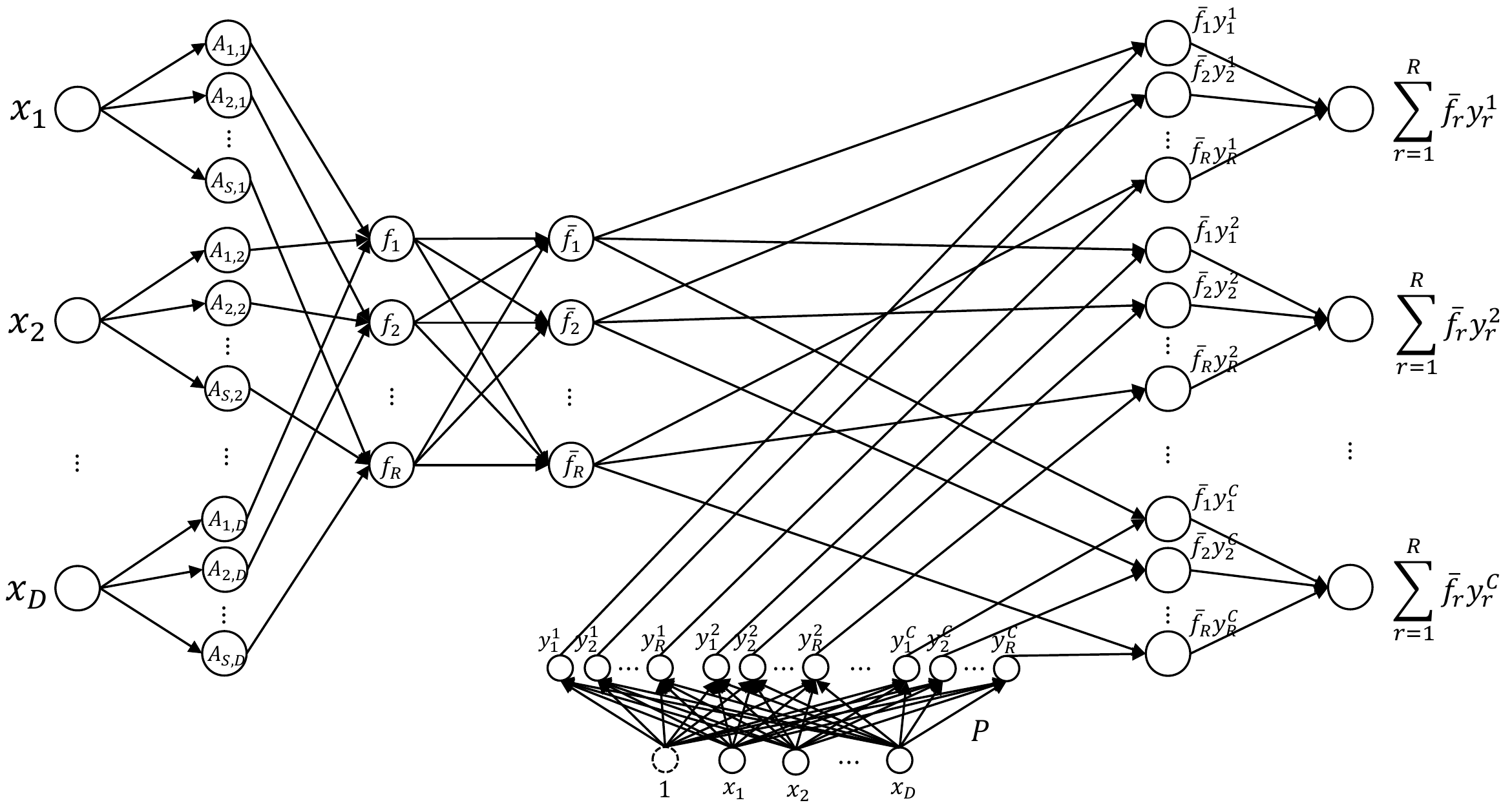}
		\caption{CoCo-FRB based TSK fuzzy system implemented by neural networks.}
		\label{compact_TSK_NNs}
	\end{figure*}

	\subsection{CoCo-FRB based TSK Fuzzy System}\label{compactTSK}
	Consider a classification problem involving $C$ classes. Let an instance or data point be represented by a $D$-dimensional feature vector, i.e., $\mathbf{x}=(x_1,x_2,\cdots,x_D)^T \in \mathbb{R}^D$. We assume that each feature is associated with $S$ fuzzy sets. For a CoCo-FRB based TSK Fuzzy System, the number of fuzzy rules, $R$, is equal to $S$ \cite{2008PalSimultaneous,2019GaoConjugate,2020CuiOptimize}. Generally, the $r_{th}~(r=1,2,\cdots,R)$ fuzzy rule of the first-order TSK model for classification problems is described as below:
	\begin{equation}\label{compactrule}
		\begin{split}
			{\rm Rule}_r:\ &{\rm IF} ~ x_1 ~ {\rm is} ~ A_{r,1} ~ {\rm and} ~ \cdots ~ {\rm and} ~ x_D ~ {\rm is} ~ A_{r,D},\\
			&{\rm THEN} ~ y_r^1(\mathbf{x})=p_{r,0}^{1}+\sum_{d=1}^{D}p_{r,d}^{1}x_d , ~ \cdots ~ , \\
			&\qquad \quad ~ y_r^C(\mathbf{x})=p_{r,0}^{C}+\sum_{d=1}^{D}p_{r,d}^{C}x_d,
		\end{split}
	\end{equation}
	where $A_{r,d}~(d=1,2,\cdots,D)$ is the fuzzy set associated with $d_{th}$ feature in the $r_{th}$ rule, $y_r^c(\mathbf{x})~(c=1,2,\cdots,C)$ means the output of the $r_{th}$ rule for the $c_{th}$ class computed from $\mathbf{x}$ and $p^{c}_{r,d}$ represents the consequent parameter of the $r_{th}$ rule associated with the $d_{th}$ feature for the $c_{th}$ class.
	
	There are plenty of choices of membership functions (MFs) that can be used for $A_{r,d}$, such as triangular, trapezoidal, and Gaussian functions \cite{2012PalAnIntegrated,2018PalFeature}. We note that the Gaussian MF is frequently employed due to its smooth and infinitely differentiable properties. Specifically, the membership value of $x_d$ on $A_{r,d}$ is:
	\begin{equation}\label{membershipvalue_old}
		\mu_{r,d}(\mathbf{x})=e^{-\frac{(x_d-m_{r,d})^2}{2\sigma_{r,d}^{2}}},
	\end{equation}
	where $m_{r,d}$ and $\sigma_{r,d}$ represent respectively the center and spread of the $r_{th}$ Gaussian membership function defined on the $d_{th}$ input variable. Note that, although in equation (\ref{membershipvalue_old}) the argument of $\mu_{r,d}$ is shown as $\mathbf{x}$, the function only uses the $d_{th}$ component of $\mathbf{x}$. In \cite{8432091}, the authors used $\mu=e^{-\frac{(x-m)^2}{\sigma^{2}}}$ as membership function but set $\sigma=1$ for all fuzzy subsystems which means \eqref{membershipvalue_old} is simplified as:
	\begin{equation}\label{membershipvalue}
		\mu_{r,d}(\mathbf{x})=e^{-(x_d-m_{r,d})^2}.
	\end{equation}
	Following \cite{8432091}, the membership values are evaluated by \eqref{membershipvalue} in this paper. Clearly, these values are all  between $0$ and $1$, i.e., $0 < \mu_{r,d}(\mathbf{x}) \le 1$.
	
	Since minimum is not differentiable, the product is often used to compute the firing strength \cite{2001PalIntegrated,2012PalAnIntegrated,2018PalFeature,2020CuiOptimize,2020WuOptimize,2020GuA,2021GuFast}:
	\begin{equation}\label{firingstrength_prod}
		f_r(\mathbf{x})=\prod_{d=1}^{D}\mu_{r,d}(\mathbf{x}),
	\end{equation}
	where $f_r(\mathbf{x})$ is the firing strength of the $r_{th}$ rule for the input $\mathbf{x}$. Note that the product T-norm often leads to a fatal flaw when the dimension, $D$, is pretty large. It easily makes $f_r(\mathbf{x})$ very close to $0$ even beyond the scope that machines can identify, that is, ``numeric underflow'' \cite{2020CuiOptimize}.
	
	The output of the $r_{th}$ rule associated the $c_{th}$ class computed from $\mathbf{x}$ is:
	\begin{equation}\label{consequentoutput}
		y_r^c(\mathbf{x})=p_{r,0}^{c}+\sum_{d=1}^{D}p_{r,d}^{c}x_d.
	\end{equation}
	If the normalized firing strength is defined as:
	\begin{equation}\label{normalizedfiringstrength}
		\bar{f_r}(\mathbf{x})=\frac{f_r(\mathbf{x})}{\sum_{i=1}^{R}f_i(\mathbf{x})},
	\end{equation}
	then the $c_{th}$ component of the system output on $\mathbf{x}$ is:
	\begin{equation}\label{modeloutput}
		y^c(\mathbf{x})=\sum_{r=1}^{R}\bar{f_r}(\mathbf{x})y_r^c(\mathbf{x}).
	\end{equation}
	
	For the sake of clarity, the neural network structure implementing CoCo-FRB based TSK fuzzy system described above is shown in Fig. \ref{compact_TSK_NNs}. By contrast, FuCo-FRB considers all possible valid combinations of fuzzy sets defined on each feature, in which the number of rules exponentially increases with the dimension of the feature space. The details of these two fuzzy rule bases will be illustrated and compared in Subsection \ref{En_FRB}.

	Assume that there are $N$ training samples. There are several choices of error functions \cite{2012PalAnIntegrated,2020WuOptimize,2020CuiOptimize} that can be used to update system parameters based on back-propagation procedure or gradient descent (GD) algorithm \cite{1986RumelhartLearning,1992WangBack,2012BengioPractical}. The typical mean square error function is used in this paper:
	\begin{equation}\label{errorfunction_batch}
		E=\frac{1}{2N}\sum_{n=1}^{N}\sum_{c=1}^{C}(y^c(\mathbf{x}_n)-y_c(\mathbf{x}_n))^2,
	\end{equation}
	where $y^c(\mathbf{x}_n)$ and $y_c(\mathbf{x}_n)$ respectively correspond to the $c_{th}$ component of the system output and the true label vector for the $n_{th}$ input instance, $\mathbf{x}_n~(n=1,2,\cdots,N)$.
	
	The gradients of the error function \eqref{errorfunction_batch} with respect to the antecedent and consequent parameters are presented as:
	\begin{equation}\label{del_mean}
		\small
		\begin{split}
			\frac{\partial E}{\partial m_{r,d}}\!=\!&\frac{1}{N}\sum_{n=1}^{N}\Bigg[ 2f_r(\mathbf{x}_n)(x_{n,d}-m_{r,d}) \\
			&\times \frac{\sum_{c=1}^{C}[(y^c(\mathbf{x}_n)-y_c(\mathbf{x}_n))(y_r^c(\mathbf{x}_n)-y^c(\mathbf{x}_n))]}{\sum_{i=1}^{R}f_i(\mathbf{x}_n)}\Bigg],
		\end{split}
	\end{equation}
	\begin{equation}\label{del_p}
		\frac{\partial E}{\partial p_{r,d}^c}=\frac{1}{N}\sum_{n=1}^{N}\left[[y^c(\mathbf{x}_n)-y_c(\mathbf{x}_n)]\bar{f_r}(\mathbf{x}_n)x_{n,d}\right],
	\end{equation}
	where $x_{n,d}$ is the $d_{th}$ component of sample $\mathbf{x}_n$. Then, the update formula of system parameters in the $k_{th}$ iteration is as follows:
	\begin{equation}\label{update}
		\omega^{(k+1)}=\omega^{(k)}-\eta \frac{\partial E}{\partial \omega^{(k)}},
	\end{equation}
	where $\omega$ indicates the general parameters of antecedent and consequent parts, $\eta>0$ is the learning rate.

	\subsection{Adaptive Softmin (Ada-softmin)}\label{ada_softmin}
	It is easy to know that the firing strength evaluated by the product T-norm takes a very small positive value even for problems with a moderate number of features. Although this issue is alleviated by using \eqref{normalizedfiringstrength} to some extent, the mentioned ``numeric underflow'' still occurs when the dimension of the problem is large enough. To deal with this issue and also to overcome the nondifferentiability of minimum, the following softmin as a substitute of minimum is often used to compute the firing strength \cite{2004PalA,2008PalSimultaneous}. 
	\begin{equation}\label{softmin}
		softmin(v_1,\!v_2,\!\cdots,\!v_D,\!q)\!=\!\left(\frac{v_1^q+v_2^q+\cdots+v_D^q}{D}\right)^\frac{1}{q}.
	\end{equation}
	As $q\rightarrow -\infty$, the softmin tends to the minimum of $v_i~(i=1,2,\cdots,D)$. Practically, the parameter $q$ is often set to be a fixed constant such as $-12$ in \cite{2004PalA} and $-11$ in \cite{2008PalSimultaneous}. Unfortunately, the softmin with fixed parameter has two drawbacks: 1) numeric underflow and 2) fake minimum. Similar to \cite{2004PalA}, we set $q = -12$ and show the following two examples to demonstrate the limitations of softmin.
	
	\textbf{Numeric underflow}: Assume that
	\begin{equation}\label{case1}
		u_1=1.1\times 10^{-26}, u_2=1.8\times 10^{-22}, u_3=1.5\times 10^{-9}.
	\end{equation}
	According to \eqref{softmin}, theoretically, $u_1^{-12}$ should be equal to $0.3186\times 10^{312}$. However, this is significantly beyond the computation scope of the machine (We are using a 64 bit based computer). The evaluation of $u_1^{-12}$ is actually shown as $\rm{Inf}$\footnote{From this point of view, the described issue can also be called ``numeric overflow'' since the fact that $u_1^{-12}=0.3186\times 10^{312}$.} during running the program code. This then leads to the phenomenon ``numeric underflow'', that is,
	\begin{equation*}
		softmin(u_1,u_2,u_3,-12)=0.
	\end{equation*}
	
	\textbf{Fake minimum}: Assume that
	\begin{equation}\label{case2}
		v_1=0.5, v_2=0.55, v_3=0.49, v_4=0.48,
	\end{equation}
	the simulation result is with
	\begin{equation*}
		softmin(v_1,v_2,v_3,v_4,-12) = 0.4977.
	\end{equation*}
	It is easy to see that softmin fails to get the minimum, $0.48$, and computes a value even  bigger than the second minimum, $v_3 = 0.49$.

	Now, we introduce an adaptive softmin called as Ada-softmin to effectively solve the above two problems. Based on the current membership values, the parameter, $q$, is adaptively assigned a suitable negative integer, $\hat{q}$, instead of a fixed value. Specifically, the firing strength is calculated with:

	\begin{equation}\label{firingstrength_adap_softmin}
		f_r(\mathbf{x})=\left(\frac{\mu_{r,1}^{\hat{q}}(\mathbf{x})+\mu_{r,2}^{\hat{q}}(\mathbf{x})+\cdots+\mu_{r,D}^{\hat{q}}(\mathbf{x})}{D}\right)^\frac{1}{{\hat{q}}},
	\end{equation}
	where
	\begin{equation}\label{qhat}
		\hat{q}=\left\lceil \frac{690}{\ln(\min\{\mu_{r,1}(\mathbf{x}),\mu_{r,2}(\mathbf{x}),\cdots,\mu_{r,D}(\mathbf{x})\})} \right\rceil,
	\end{equation}
    and $\left\lceil\cdot \right\rceil$ is the ceiling function. We note that the adaptive parameter, $\hat{q}$, is deduced by the assumption: each $\mu_{r,d}^{q}(\mathbf{x})$ is not greater than $10^{300}$. This clearly guarantees that the value of $\mu_{r,d}^{q}(\mathbf{x})$ lies in the scope of a 64 bit machine. By the proposed Ada-softmin, one can easily obtain the new results on \eqref{case1} and \eqref{case2}:
	\begin{equation}\label{case1_res}
		\left\{
		\begin{split}
			&\hat{q}_2=\left\lceil \frac{690}{\ln(\min\{u_1,u_2,u_3\})} \right\rceil=-11,\\
			&softmin(u_1,u_2,u_3,\hat{q}_2)=1.2\times 10^{-26}.
		\end{split}
		\right.
	\end{equation}
	\begin{equation}\label{case2_res}
		\left\{
		\begin{split}
			&\hat{q}_1=\left\lceil \frac{690}{\ln(\min\{v_1,v_2,v_3,v_4\})} \right\rceil=-940,\\
			&softmin(v_1,v_2,v_3,v_4,\hat{q}_1)=0.4807,
		\end{split}
		\right.
	\end{equation}
	As shown in \eqref{case1_res} and \eqref{case2_res}, the proposed Ada-softmin can get better approximations to the minimum in these two cases. To further avoid the ``numeric underflow'' problem, we set $-1,000$ as the lower bound of $\hat{q}$. Specifically, if the $\hat{q}$ calculated from \eqref{qhat} is less than $-1,000$, then we set this $\hat{q}$ to $-1,000$. Since the proposed Ada-softmin has the ability to adaptively find the appropriate parameter of the softmin, the AdaTSK can be successfully used for the high-dimensional problems. The classification results of TSK fuzzy systems with product, softmin and Ada-softmin are compared in Subsection \ref{AdaTSK_classifier}, which demonstrate the effectiveness of AdaTSK.

	\section{Feature Selection and Rule Extraction Based on AdaTSK (FSRE-AdaTSK)}\label{FSRE_AdaTSK_method}
	In this section, we introduce a new gate function which is embedded in AdaTSK fuzzy system for feature selection and rule extraction. For feature selection, the fuzzy rule base, CoCo-FRB, is considered. While for rule extraction, an enhanced fuzzy rule base called as En-FRB is designed inspired by CoCo-FRB and FuCo-FRB. To some extent, En-FRB can be regarded as a compromised solution between them.

	\subsection{Gate Function}\label{gate_function_subsection}
	\begin{figure}[t]
		\centering
		\subfigure[Gate functions]{\includegraphics[scale=0.6]{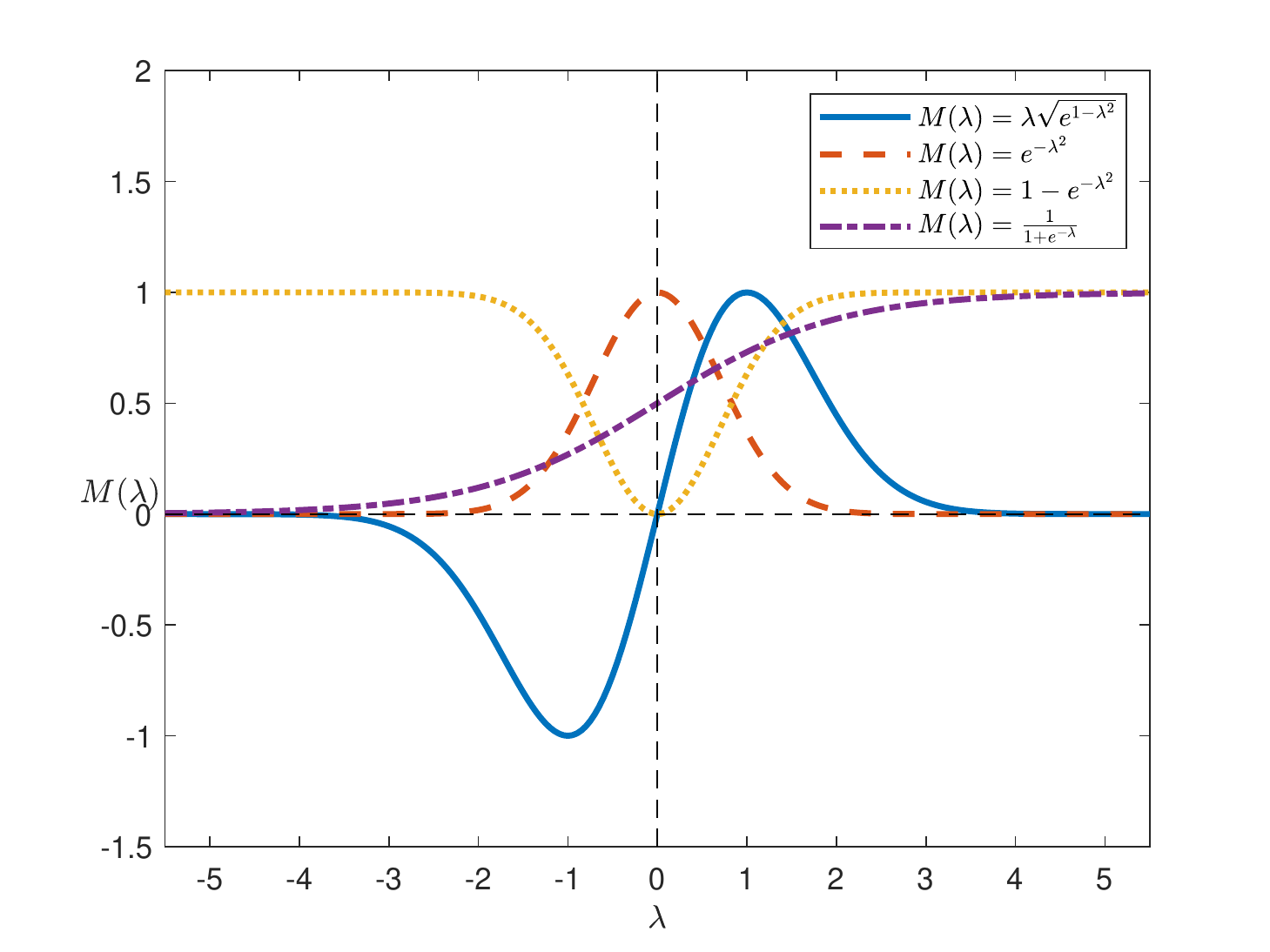}\label{gatefun}}
		\vfil
		\subfigure[Derivatives of gate functions] {\includegraphics[scale=0.6]{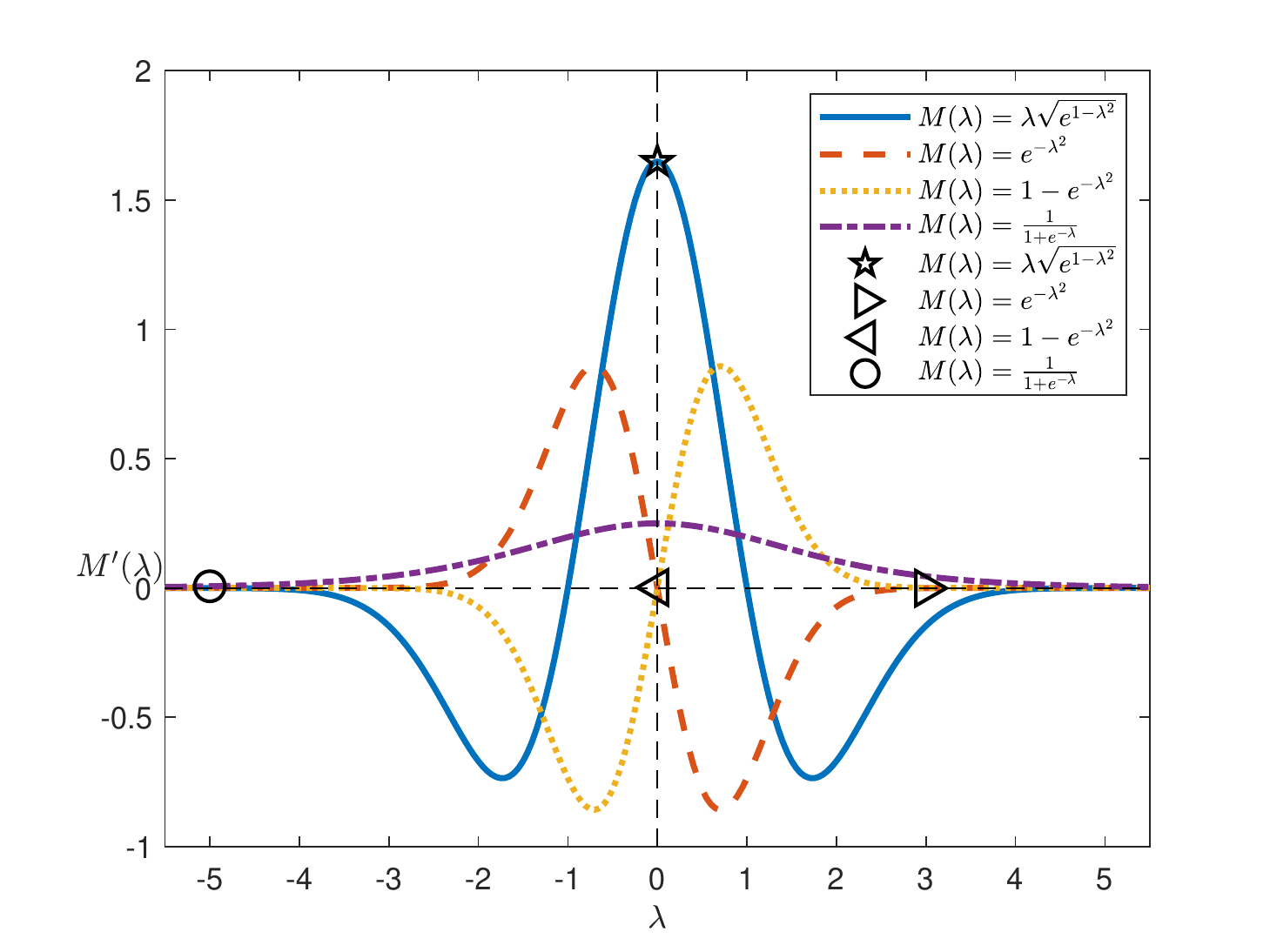}\label{gatefun_derivative}}
		\caption{Gate functions and their derivatives}
		\label{gate_plot}
	\end{figure}
	
	Each T-norm, $T$, satisfies the equation $T(1,\mu)=\mu$, which means $1$ has no effect on the result of T-norms. In a fuzzy rule based system, if we can make the membership value of a feature is always $1$ irrespective of the actual numerical value of the feature, then this feature makes no difference to the firing strength. Some of the earlier studies used a gate function to modulate membership value of each feature in such way that the features whose modulated membership value is $1$ can be removed from the feature space and rule base. For example, assume that the membership value $\widetilde{\mu}$ is calculated from the feature $\widetilde{F}$ using \eqref{membershipvalue}. The gate function $M(\lambda)$ can modulate $\widetilde{\mu}$ as $\widetilde{\mu}^{M(\lambda)}$. We can then use gradient descent based algorithm to learn the gate parameter $\lambda$. When $M(\lambda)=0$ the feature $\widetilde{F}$ has no contribution to the firing strength and consequently is pruned. Every feature is equipped with a gate function and the features with high gate values are important features which are selected \cite{2001PalIntegrated,2004PalA,2008PalSelecting,2012PalAnIntegrated,2018PalFeature}. In addition, for the first-order TSK fuzzy model, the gate function needs to be embedded on both antecedent and consequent parts for feature selection \cite{2008PalSimultaneous}.  

	{\bf Note:} In this paper we take a completely different philosophy. We do not modify the antecedent membership values using gate function as done in \cite{2001PalIntegrated,2004PalA,2008PalSelecting,2012PalAnIntegrated,2018PalFeature,2008PalSimultaneous}. We use  two families of gate functions. The first family ($M(\lambda)$) is used for feature selection, while the second family ($M(\theta)$) is used for rule extraction. These two families of gate functions are used in two different phases of the training and in both cases the gates are used {\it only} with the consequents.
	
	In our proposed approach, following \cite{2004PalA,2012PalAnIntegrated,2015PalFeature}, at the beginning, all features are regarded as poor features, i.e., the gates, $M(\lambda)$s, of the features are initialized to low values. However, we note that for the commonly used gate functions, such as $M(\lambda)=\frac{1}{1+e^{-\lambda}}$, $M(\lambda)=1-e^{-\lambda^2}$ and $M(\lambda)=e^{-\lambda^2}$, their respective derivatives are close to zero, when the gate values are close to zero. This makes the learning very slow at the beginning. Motivated by this observation, we propose the following gate function: 
	\begin{equation}\label{new_gate_function}
		M(\lambda)=\lambda\sqrt{e^{1-\lambda^2}},
	\end{equation}
    where $\lambda$ is the gate parameter.
    
    In Fig. \ref{gatefun}, the curves of the above four gate functions are plotted. Since initially all gates are assumed to be closed, the corresponding parameters, $\lambda$s, are located in different regions. For $M(\lambda)=\frac{1}{1+e^{-\lambda}}$ used in \cite{2015PalFeature}, the $\lambda$s are initialized with $-5.0\pm$ random noise in $[0,1]$. For $M(\lambda)=1-e^{-\lambda^2}$ \cite{2004PalA} and $M(\lambda)=e^{-\lambda^2}$ \cite{2012PalAnIntegrated}, the $\lambda$s are respectively initialized with $0.001$ and $3.0+\mathcal{N}(0,0.2)$ where $\mathcal{N}(0,0.2)$ indicates the Gaussian noise with mean zero and standard deviation $0.2$. For the proposed gate function, \eqref{new_gate_function}, the parameter values to indicate a closed gate are located in three different zones: around zero, around $-3.0$, and around $+3.0$.  Besides, it is easy to see that the gate function, \eqref{new_gate_function}, is an odd function. It is significantly different from the other three gate functions which lie always above the horizontal axis. As we can see from Fig. \ref{gatefun}, $M(\lambda)$ could be negative. As far as the consequent is concerned, a high magnitude negative gate value will indicate a useful feature. However, features with absolute gate value close to zero can be deleted. Thus, the proposed gate function can be viewed as a bidirectional door, one can open it by pushing it or pulling it with an opposite direction. In other words, for \eqref{new_gate_function}, its absolute value can be denoted as the opening degree of gate. 
          
	The derivatives of the above four gate functions are graphed in Fig. \ref{gatefun_derivative}. For almost closed initial gates, one can notice that the magnitude of the derivative of the proposed gate function, \eqref{new_gate_function}, is much greater than those of its counterparts when its parameter, $\lambda$, is located around zero. This, thus, enables the proposed gate function to learn to distinguish between good and poor features faster compared to the other choices. As an example, comparing with $M(\lambda)=e^{-\lambda^2}$, we do a simulation on Wine dataset to verify the advantage of the proposed gate function, \eqref{new_gate_function}. The details are presented in Subsection \ref{gate_function_comparison}.

	\begin{figure}[t]
		\centering
		\subfigure[CoCo-FRB]{\includegraphics[scale=0.4]{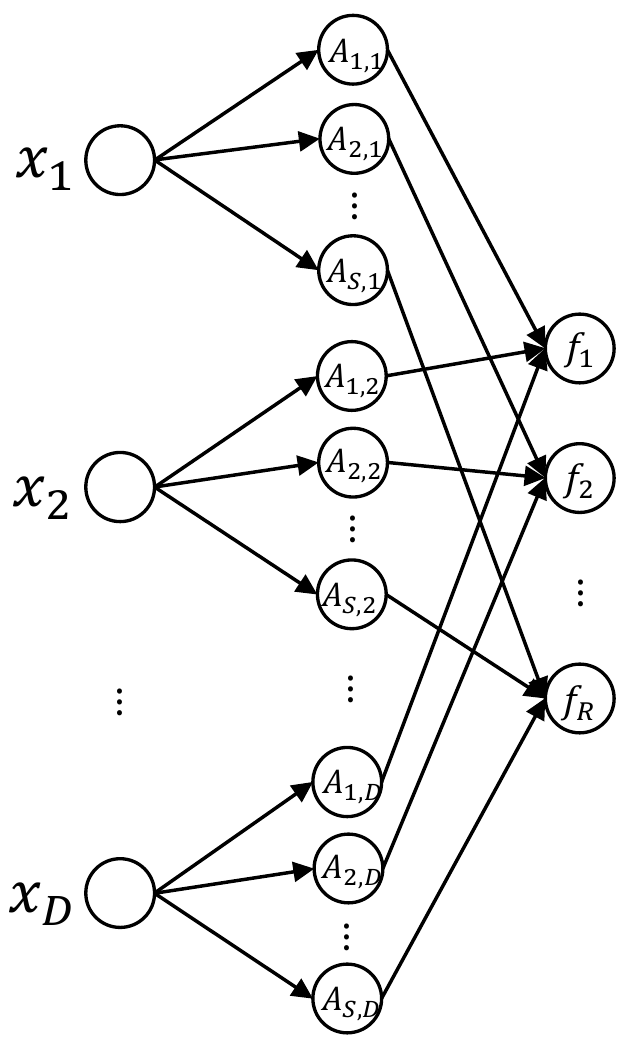}\label{CoCo-FRB-structure}}
		\hfil
		\subfigure[FuCo-FRB]{\includegraphics[scale=0.4]{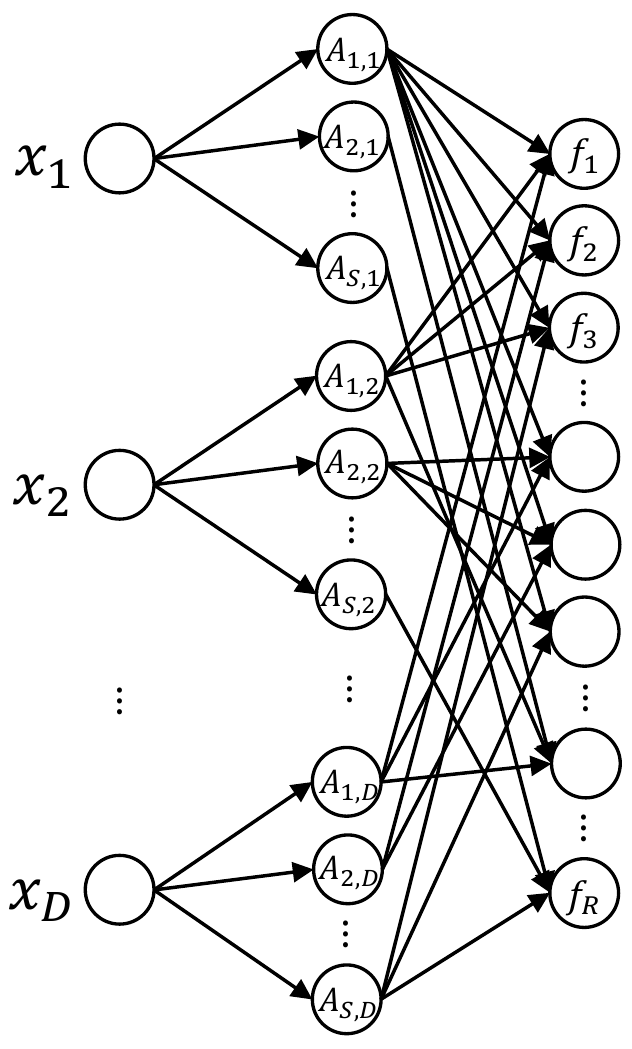}\label{FuCo-FRB-structure}}
		\caption{Neural network structure of antecedent parts in CoCo-FRB and FuCo-FRB}
		\label{CoCo-FRB-FuCo-FRB-structure}
	\end{figure}

	\subsection{Enhanced Fuzzy Rule Base (En-FRB)}\label{En_FRB}
	In this section, the enhanced fuzzy rule base, En-FRB, is constructed which is an intermediate case of CoCo-FRB and FuCo-FRB. The antecedent parts in CoCo-FRB and FuCo-FRB are shown with two different network architectures in Fig. \ref{CoCo-FRB-FuCo-FRB-structure}.
	The number of rules in these two fuzzy rule bases are $S$ and $S^D$, respectively. FuCo-FRB considers all possible feasible rules in which all fuzzy sets corresponding to every feature are combined with each other, that is, $R=S^D$. Clearly, FuCo-FRB is not a feasible approach for high-dimensional problems. While CoCo-FRB could be regarded as a simplified case of FuCo-FRB, $R=S$. None of these two types of fuzzy rule bases may be suitable for rule extraction because CoCo-FRB may not be expressive enough while FuCo-FRB is too big to consider. However, if we could choose the right number of clusters CoCo-FRB may solve the problem, but choosing the right number of clusters is a difficult problem.
	
\begin{figure*}[ht] 		
		\centering
		\includegraphics[scale=0.5]{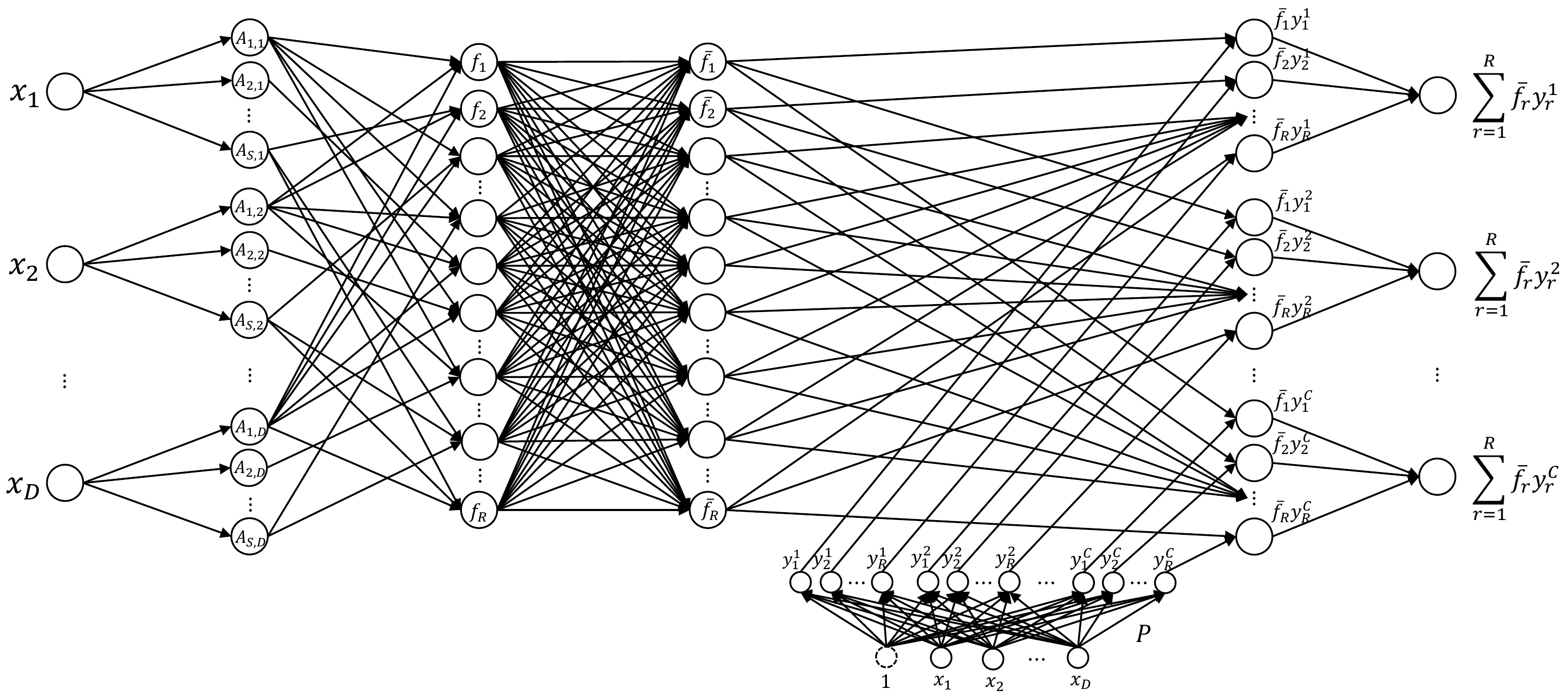}
		\caption{En-FRB based TSK fuzzy system implemented by neural networks.}
		\label{enhanced_TSK_NNs}
	\end{figure*}

	To construct the En-FRB, we first introduce the following index matrix (I-matrix) to represent the combination of fuzzy sets. For CoCo-FRB shown in Fig. \ref{CoCo-FRB-structure}, the I-matrix is
	\begin{equation*}
		I_{CoCo}=\left[
		\begin{matrix}
			1 &1 &\cdots &1\\
			2 &2 &\cdots &2\\
			\vdots &\vdots &\ddots &\vdots\\
			S &S &\cdots &S
		\end{matrix}\right]_{S\times D},
	\end{equation*}
	where $I_{CoCo}(i,j)~(i=1,2,\cdots,S;~j=1,2,\cdots,D)$ is the index of the fuzzy set associated with $j_{th}$ feature in $i_{th}$ rule. For example, $I_{CoCo}(2,1)=2$ means the second fuzzy set associated with the first feature is used in the second rule. As for FuCo-FRB, the I-matrix is
	\begin{equation*}
		I_{FuCo}=\left[
		\begin{matrix}
			1 &1 &\cdots &1\\
			1 &1 &\cdots &2\\
			\vdots &\vdots &\ddots &\vdots\\
			1 &1 &\cdots &S\\
			\vdots &\vdots &\ddots &\vdots\\
			S &S &\cdots &1\\
			\vdots &\vdots &\ddots &\vdots\\
			S &S &\cdots &S
		\end{matrix}\right]_{S^D\times D}.
	\end{equation*}
	To extend CoCo-FRB and also avoid the exponential increase of number of rules for FuCo-FRB, the I-matrix of En-FRB is designed as follows:
	\begin{equation*}
		I_{En}=\left[
		\begin{matrix}
			1 &1 &\cdots &1\\
			S &1 &\cdots &1\\
			\vdots &\vdots &\ddots &\vdots\\
			1 &1 &\cdots &S\\
			2 &1 &\cdots &1\\
			\vdots &\vdots &\ddots &\vdots\\
			1 &1 &\cdots &2\\
			\vdots &\vdots &\ddots &\vdots\\
			S &S &\cdots &S\\
			S-1 &S &\cdots &S\\
			\vdots &\vdots &\ddots &\vdots\\
			S &S &\cdots &S-1\\
			1 &S &\cdots &S\\
			\vdots &\vdots &\ddots &\vdots\\
			S &S &\cdots &1
		\end{matrix}\right]_{(2D+1)S\times D}.
	\end{equation*}
	The En-FRB extends $2\times D$ rules from each rule in CoCo-FRB and thus, $2DS+S=(2D+1)S$ rules are obtained. As a consequence, the number of rules $R$ in En-FRB increases linearly rather than exponentially with the number of features $D$. Based on En-FRB, the rule extraction can be done using the TSK fuzzy system. For convenience, we use $I$ to represent the I-matrix of En-FRB, then the $r_{th}$ rule in En-FRB associated with a specific sample, $\mathbf{x} = (x_1, x_2, \cdots, x_D)^T$, could be rewritten as:
	\begin{equation}\label{enhancedrule}
		\begin{split}
			{\rm Rule}_r: \!~ &{\rm IF} \!~ x_1 \!~ {\rm is} \!~ A_{I(r,1),1} \!~ {\rm and} \!~ \cdots \!~ {\rm and} \!~ x_D \!~ {\rm is} \!~ A_{I(r,D),D},\\
			&{\rm THEN} ~ y_r^1(\mathbf{x})=p_{r,0}^{1}+\sum_{d=1}^{D}p_{r,d}^{1}x_d, ~ \cdots ~ ,\\
			&\qquad \quad ~ y_r^C(\mathbf{x})=p_{r,0}^{C}+\sum_{d=1}^{D}p_{r,d}^{C}x_d,
		\end{split}
	\end{equation}
	where $r=1,2,\cdots,R$ and $R=(2D+1)S$. The neural network structure of En-FRB based TSK fuzzy system is shown in Fig. \ref{enhanced_TSK_NNs}.

	\subsection{Feature Selection and Rule Extraction}\label{FSRE_method}
\begin{figure}[t]
		\centering
		\subfigure[Gates for feature selection]	{\includegraphics[scale=0.54]{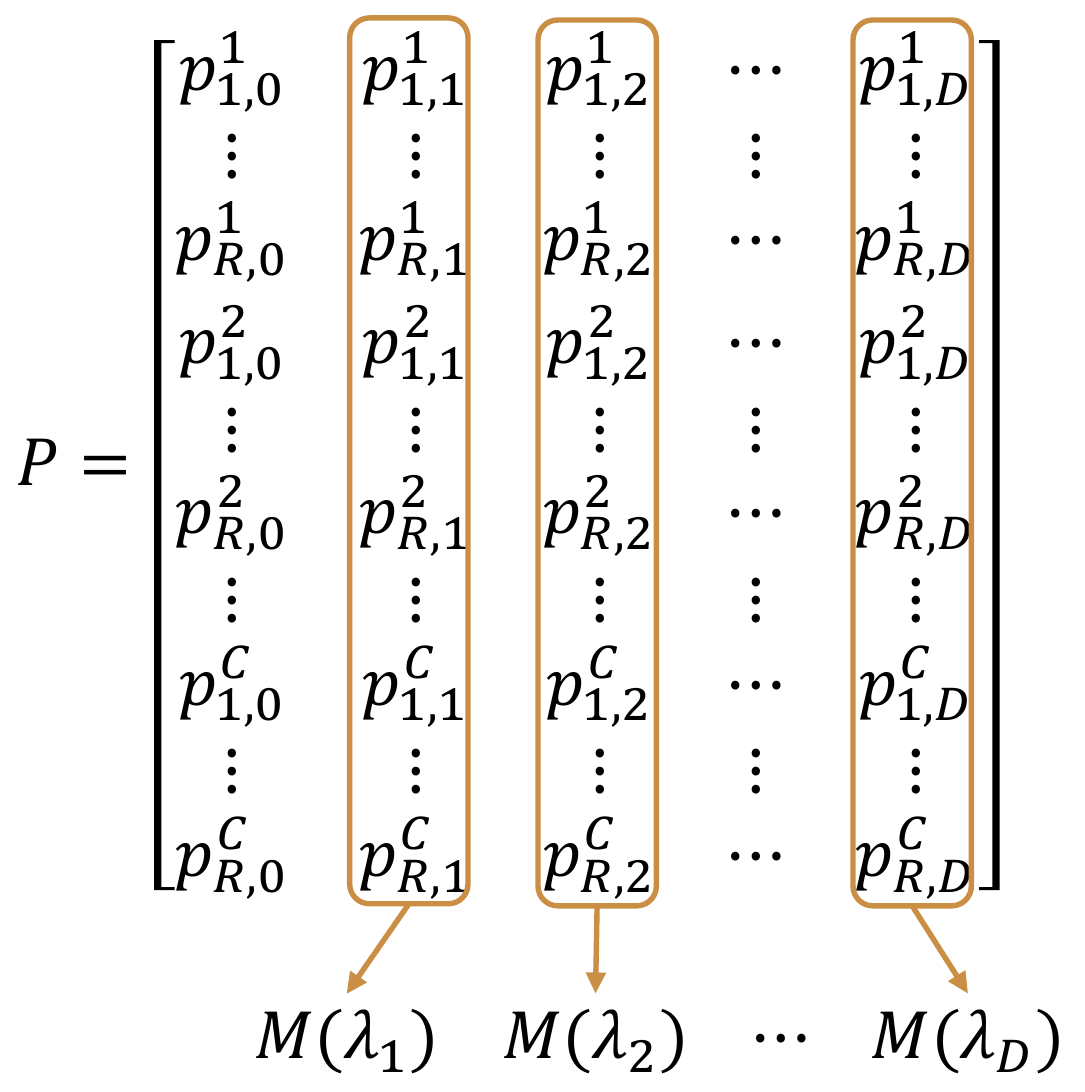}\label{gate_FS}}
		\vfil
		\subfigure[Gates for rule extraction] {\includegraphics[scale=0.54]{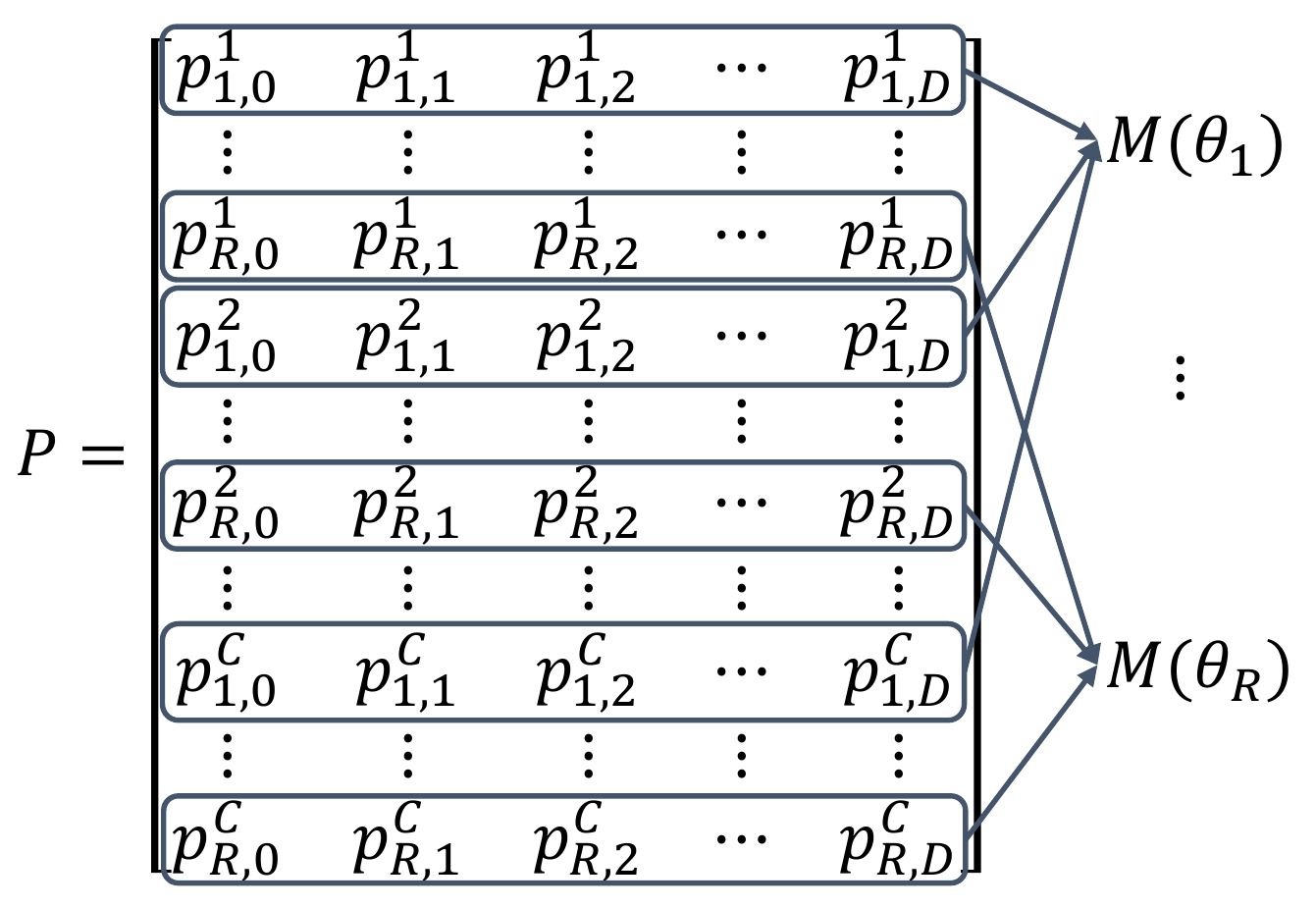}\label{gate_RE}}
		\caption{Using gate function to multiply consequent parameters for feature selection and rule extraction}
		\label{gate_FS_RE}
	\end{figure}

	To improve the classification performance, the proposed gate function, \eqref{new_gate_function}, is used for feature selection and rule extraction on the constructed AdaTSK system. For feature selection, the consequent parameters associated with the $d_{th}$ feature are multiplied by the corresponding gate function $M(\lambda_d)$, thus, \eqref{consequentoutput} is modified as:
	\begin{equation}\label{consequentoutput_FS}
		y_r^c(\mathbf{x})=p_{r,0}^{c}+\sum_{d=1}^{D}M(\lambda_d)p_{r,d}^{c}x_d.
	\end{equation}
	On the other hand, for rule extraction, the entire consequent, i.e., all consequent parameters in the $r_{th}$ rule are multiplied by the rule-gate function $M(\theta_r)$. Thus, \eqref{consequentoutput} is modified as:
	\begin{equation}\label{consequentoutput_RE}
		y_r^c(\mathbf{x})=M(\theta_r)p_{r,0}^{c}+M(\theta_r)\sum_{d=1}^{D}p_{r,d}^{c}x_d,
	\end{equation}
	where $\lambda_d$ and $\theta_r$ are updated along with the system parameters. Note that these two families of gates are used in two sequential phases of training, not simultaneously. Fig. \ref{gate_FS} and Fig. \ref{gate_RE} indicate the sets of parameters, which are multiplied by gate function for feature selection and gate function for rule extraction, respectively.
	
	As mentioned earlier, the final FSRE-AdaTSK scheme includes three sequential phases: (i) feature selection, (ii) rule extraction and (iii) fine tuning. Specifically, in phase (i), we do the feature selection based on CoCo-FRB. And in phase (ii), the En-FRB is constructed based on the reduced feature space, the rule extraction is then implemented on it. Note that the rule extraction phase also discards rules that are not useful. In phase (iii), the system parameters are fine-tuned to improve the classification performance. The proposed gate function is used in both phase (i) and phase (ii) and not in phase (iii). The Ada-softmin is used in all of these three phases.
	
	For feature selection, the gradient of error function with respect to the consequent parameters is rewritten as:
	\begin{equation}\label{del_p_gate_FS}
		\small
		\frac{\partial E}{\partial p_{r,d}^c}\!=\!\frac{1}{N}\sum_{n=1}^{N}\left[[y^c(\mathbf{x}_n)-y_c(\mathbf{x}_n)]\bar{f_r}(\mathbf{x}_n)M(\lambda_d)x_{n,d}\right],
	\end{equation}
	On the other hand, for rule extraction, the gradient of error function with respect to the consequent parameters is derived as:
	\begin{equation}\label{del_p_gate_RE}
		\small
		\frac{\partial E}{\partial p_{r,d}^c}=\frac{1}{N}\sum_{n=1}^{N}\left[[y^c(\mathbf{x}_n)-y_c(\mathbf{x}_n)]\bar{f_r}(\mathbf{x}_n)M(\theta_r)x_{n,d}\right].
	\end{equation}
	While in phase (iii), since already the feature and the rules have been selected and the gates have been removed, the gradient of error function with respect to the consequent parameters is the same as in \eqref{del_p}. Since the gate function has no effect on the antecedent part, the formula to compute the gradients of error function with respect to centers are the same in these three phases, which is presented in \eqref{del_mean_adapsoftmin}. The equations \eqref{del_lambda} and \eqref{del_theta} show the gradients of the error function with respect to $\lambda$s and $\theta$s, respectively. All of these parameters are updated by the general update rule in \eqref{update}.
	\begin{figure*}[t]
		\begin{equation}\label{del_mean_adapsoftmin}
			\frac{\partial E}{\partial m_{r,d}}=\frac{1}{N_b}\sum_{n=1}^{N_b}\left[2f_r(\mathbf{x}_n)(x_{n,d}-m_{r,d})
			\frac{\mu_{r,d}^{\hat{q}}(\mathbf{x}_n)}{\mu_{r,1}^{\hat{q}}(\mathbf{x}_n)+\cdots+\mu_{r,D}^{\hat{q}}(\mathbf{x}_n)}
			\frac{\sum_{c=1}^{C}[(y^c(\mathbf{x}_n)-y_c(\mathbf{x}_n))(y_r^c(\mathbf{x}_n)-y^c(\mathbf{x}_n))]}{\sum_{i=1}^{R}f_i(\mathbf{x}_n)}\right].
		\end{equation}
		\begin{equation}\label{del_lambda}
			\frac{\partial E}{\partial \lambda_d}=\frac{1}{N_b}(1-\lambda_d^2)\sqrt{1-\lambda_d^2}\sum_{n=1}^{N_b}\sum_{r=1}^{R}\sum_{c=1}^{C}\left[x_{n,d}[y^c(\mathbf{x}_n)-y_c(\mathbf{x}_n)]\bar{f_r}(\mathbf{x}_n)p_{r,d}^{c}\right].
		\end{equation}
		\begin{equation}\label{del_theta}
			\frac{\partial E}{\partial \theta_r}=\frac{1}{N_b}(1-\theta_r^2)\sqrt{1-\theta_r^2}\sum_{n=1}^{N_b}\sum_{c=1}^{C}\left[\bar{f_r}(\mathbf{x}_n)[y^c(\mathbf{x}_n)-y_c(\mathbf{x}_n)](p_{r,0}^{c}+\sum_{d=1}^{D}p_{r,d}^{c})\right].
		\end{equation}
	\end{figure*}

	We use a threshold, $\tau_\lambda$, to select useful features and another threshold, $\tau_\theta$, to extract important rules. The phase (i) training starts with almost closed gates for all features and the phase (ii) training starts will almost closed gates for all rules.

	The thresholds for feature selection and rule extraction are computed as follows:
	\begin{equation}\label{tau_FS}
		\small
		\tau_{\lambda}=\max\limits_{d}\{M(\lambda_d)\}-\zeta_{\lambda}\left[\max\limits_{d}\{M(\lambda_d)\}-\min\limits_{d}\{M(\lambda_d)\}\right],
	\end{equation}
	\begin{equation}\label{tau_RE}
		\small
		\tau_{\theta}=\max\limits_{r}\{M(\theta_r)\}-\zeta_{\theta}\left[\max\limits_{r}\{M(\theta_r)\}-\min\limits_{r}\{M(\theta_r)\}\right],
	\end{equation}
	where $\zeta_{\lambda}$ and $\zeta_{\theta}$ are the coefficients to compute thresholds for feature selection and rule extraction, respectively. In this paper, we adopt different thresholds for low and high-dimensional datasets, $\zeta_{\lambda}=0.5$, $\zeta_{\theta}=0.3$ for the low-dimensional datasets and $\zeta_{\lambda}=0.4$, $\zeta_{\theta}=0.5$ for high-dimensional datasets. When the number of extracted rules is smaller than the number of classes, it may make the learning of the classifier more challenging. Thus, the lower bound on the number of the extracted rules is fixed as the number of classes, $C$. We note here that for the TSK classifier model, as we shall demonstrate in Section \ref{AdaTSK_classifier}, it is not necessary to have at least $C$  rules. We enforce this as it will facilitate the learning of the classifier. To make these three phases more clear, the steps of FSRE-AdaTSK is described in Algorithm \ref{alg-1}. 
	
	\begin{algorithm}[t]
		\caption{The FSRE-AdaTSK algorithm}
		\begin{algorithmic}[1]
			\Require
			Training set along with class lable of each instance;
			The number of  fuzzy sets on each feature along with their definitions to define the initial rule base;
			The maximum number of training iterations in each of the three phases;
			The learning rate;
			\Ensure
			The final system and gate parameters.
			\State Step 1. Initialize CoCo-FRB and the parameters of gate functions for feature selection. Calculate the forward propagation process by \eqref{membershipvalue}, \eqref{firingstrength_adap_softmin}, \eqref{normalizedfiringstrength}, \eqref{consequentoutput_FS} and \eqref{modeloutput}, and compute system error by \eqref{errorfunction_batch}. Then  compute the gradients of the error with respect to the  system parameters and gate parameters by \eqref{del_mean}, \eqref{del_p_gate_FS} and \eqref{del_lambda}, and update the parameters a given number of iterations using \eqref{update}.   
			\State Step 2. Compute threshold $\tau_{\lambda}$ for each feature by \eqref{tau_FS}. Select features with gate values more than $\tau_{\lambda}$ and simplify the rule base.
			\State Step 3. Reinitialize En-FRB using the reduced feature space as well as the parameters of gate functions for rule extraction.  Calculate the forward propagation process by \eqref{membershipvalue}, \eqref{firingstrength_adap_softmin}, \eqref{normalizedfiringstrength}, \eqref{consequentoutput_RE}, and \eqref{modeloutput}, and compute the system error by \eqref{errorfunction_batch}. Then compute the gradients of the error with respect to the  system parameters and gate parameters by \eqref{del_mean}, \eqref{del_p_gate_RE} and \eqref{del_theta}, and update the parameters a given number of iterations using \eqref{update}.   
			\State Step 4. Compute threshold $\tau_{\theta}$ for each rule by \eqref{tau_RE}. Retain only the rules with gate values more than $\tau_{\theta}$; if the number of extracted rules is less than the number of classes, then retain  $C$ rules with the highest gate values. Finally, the simplified fuzzy rule base is obtained.
			\State Step 5. Based on the reduced feature space and the simplified fuzzy rule base, calculate the forward propagation process by \eqref{membershipvalue}, \eqref{firingstrength_adap_softmin}, \eqref{normalizedfiringstrength}, \eqref{consequentoutput} and \eqref{modeloutput}, and compute system error by \eqref{errorfunction_batch}. Then compute the gradients of the system error with respect to the centers and consequent parameters by \eqref{del_mean} and \eqref{del_p}, and update them a certain number of iterations using \eqref{update}. This concludes design process of the rule base.
		\end{algorithmic}\label{alg-1}
	\end{algorithm}

	\section{Experiments and Results}\label{experiments_results}
	\begin{table}[t]
		\centering
		\caption{Summary of the 24 classification datasets}
		\begin{tabular}{c c c c}
			\hline
			\hline
			Dataset & \#Features & \#Classes & Dataset Size\\
			\hline
			Iris & 4 & 3 & 150\\
			Appendicitis & 7 & 2 & 106\\
			Pima & 8 & 2 & 768\\
			Yeast & 8 & 10 & 1484\\
			Glass & 9 & 6 & 214\\
			Page-blocks & 10 & 5 & 5473\\
			Wine & 13 & 3 & 178\\
			Heart & 13 & 2 & 270\\
			Wdbc & 30 & 2 & 569\\
			Texture & 40 & 11 & 5500\\
			Spectfheart & 44 & 2 & 267\\
			Sonar & 60 & 2 & 208\\
			\hline
			ORL & 1024 & 40 & 400\\
			Colon & 2000 & 2 & 62\\
			SRBCT & 2308 & 4 & 83\\ 
			ARP & 2400 & 10 & 130\\
			PIE & 2420 & 10 & 210\\
			Leukemia & 7129 & 2 & 72\\
			CNS & 7129 & 5 & 42\\
			\hline
			\hline
		\end{tabular}
		\label{datasets}
	\end{table}
	
	To demonstrate the effectiveness of AdaTSK, $19$ classification datasets are tested in the experiments. Table \ref{datasets} summarizes the information on these datasets, which includes the number of features (\#Features), the number of classes (\#Classes) and, the size of datasets. {\it Note that the datasets with more than $1,000$ features are considered high-dimensional problems in this investigation.}

	For our experiments, we use the following computational protocols: For each feature we define $S$ fuzzy sets, each modeled by a Gaussian membership function to have a simple fuzzy system. The centers of the membership functions are evenly placed on the domain of the feature as defined by the minimum and the maximum feature values in the training set. All consequent parameters are initialized to zero. We note here that after the first update cycle, the consequent parameters will take non-zero values. The same protocol is followed for all datasets. Then we use full batch GD algorithm to optimize this AdaTSK system. We observed that, for high-dimensional datasets, at the end of the training the centers of the memberships do not change much from their initial values. The centers are barely updated significantly regardless of whether we use product, softmin, or Ada-softmin as the T-norm. Therefore, we keep the centers fixed at their initial values when solving high-dimensional datasets. As a result of this, the appropriate values of the parameter of Ada-softmin, for each rule and each instance, $\hat{q}$s, are assigned before training starts and these values do not change with iterations since the membership values are not changed. However, for low-dimensional datasets, all the system parameters are updated during the training.

	\begin{figure*}[t] 		
		\centering
		\subfigure[$M(\lambda)=e^{-\lambda^2}$]{\includegraphics[scale=0.4]{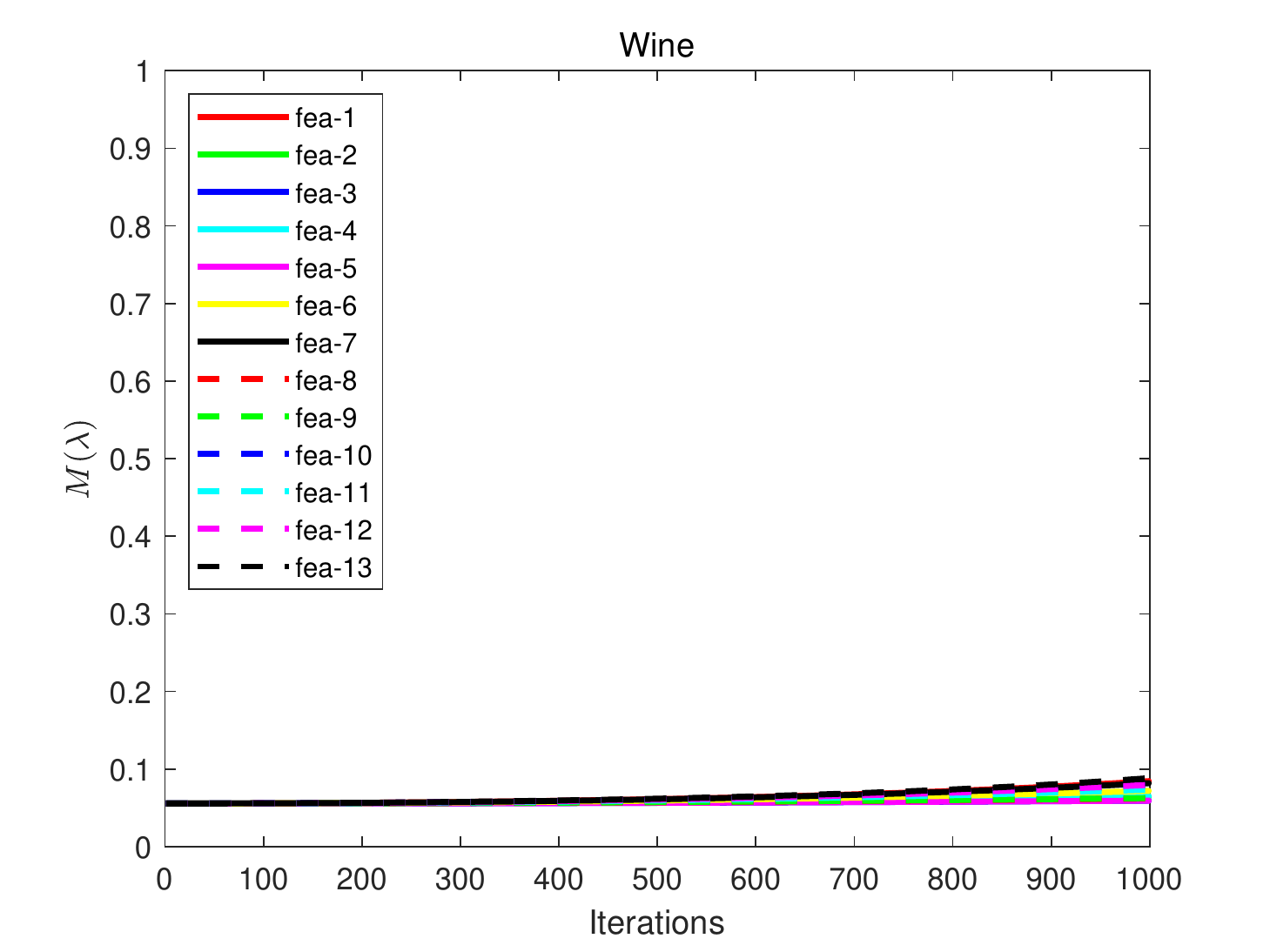}\label{Wine_FS_old_gate}}
		\subfigure[$M(\lambda)=\lambda\sqrt{e^{1-\lambda^2}}$] {\includegraphics[scale=0.4]{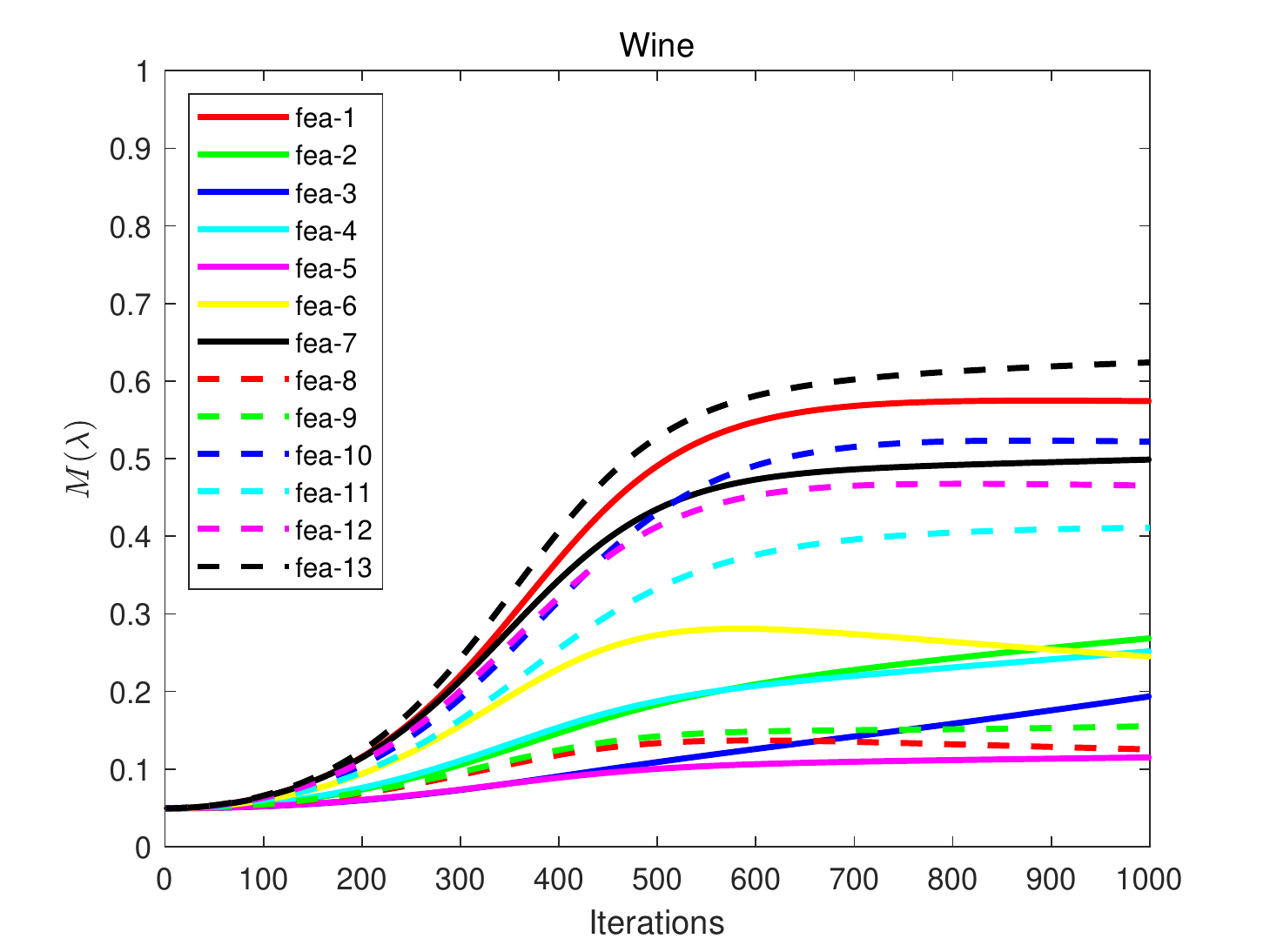}\label{Wine_FS_new_gate}}
		\subfigure[$M(\lambda)=e^{-\lambda^2}$]{\includegraphics[scale=0.4]{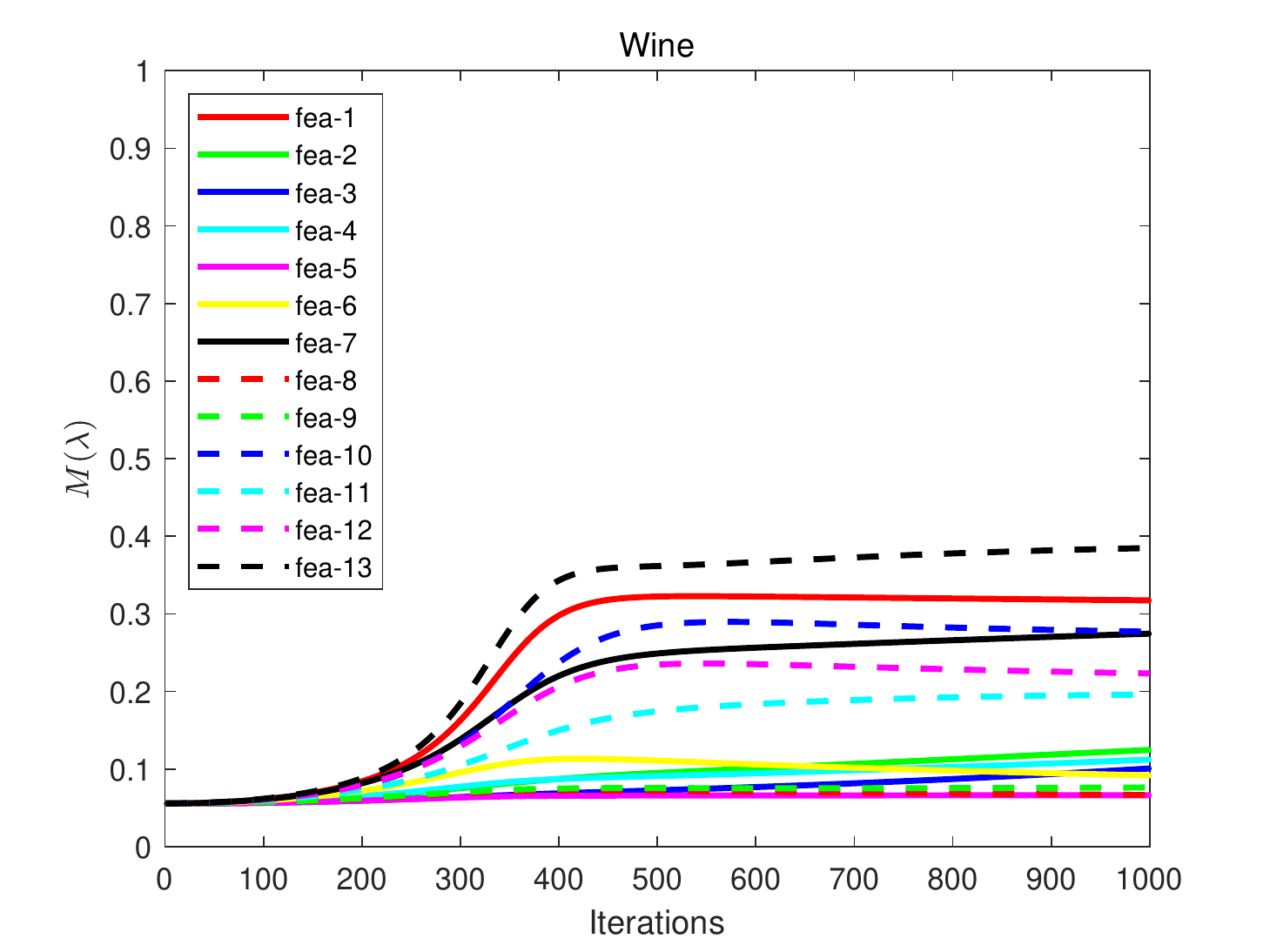}\label{Wine_FS_old_gate_lr_005}}
		\caption{The gate openings of different gate functions on the Wine dataset: For (a) and (b), the learning coefficient is $0.01$; for (c) the learning coefficient is $0.05$}
		\label{wine_gate_plot}
	\end{figure*}

	In the numerical experiments, the datasets are partitioned into the training and test sets. To minimize the effect of initial setting, ten-fold cross-validation mechanism \cite{2012PalAnIntegrated,2018PalFeature,2015PalFeature,2021WangFeature} is used for the simulations. The instances (along with their labels) of datasets are randomly divided into ten equal subsets to the extent possible, then one subset is selected as the test set and the union of the remaining nine subsets is used as the training set. The process rotates using each subset as the test set. The cross-validation experiment is repeated $10$ times and the average performance is reported. 
	
	In order to verify the novelties of the proposed approach, we conduct the following three-part experiments. In Subsection \ref{gate_function_comparison}, we first demonstrate the efficiency of the proposed gate function in learning importance of features, \eqref{new_gate_function}. For this, we use the Wine dataset, as an example. In Subsection \ref{AdaTSK_classifier}, we show the efficiency and robustness of the proposed Ada-softmin especially on high-dimensional  classification datasets. For this, we do not try to select features or reduce the size of the rule base. Finally, in Subsection \ref{FSRE_AdaTSK_results}, we demonstrate how effective  the proposed approach, FSRE-AdaTSK, is for feature selection as well as rule extraction for classification of the $19$ datasets. We also compare the performance of FSRE-AdaTSK with a state-of-the-art method on $12$ datasets.

	\subsection{Comparisons of Gate Functions}\label{gate_function_comparison}
	As mentioned in Subsection \ref{gate_function_subsection}, the proposed gate function, \eqref{new_gate_function}, can learn to distinguish between useful and poor features faster than the existing ones. To demonstrate this advantage, based on the feature selection method mentioned in Subsection \ref{FSRE_method} and AdaTSK using CoCo-FRB, the learning ability of \eqref{new_gate_function} and $M(\lambda)=e^{-\lambda^2}$ are compared on the Wine dataset.
	
	As explained in the computational protocol, we use three linguistic values, each modeled by a Gaussian  membership function, for each feature. The system parameters are also initialized as we explained earlier. The $\lambda$s are initialized in such a manner that the gate values, $M(\lambda)$s, are initialized to values near $0.05$ ($\lambda$s are around zero) to ensure that at the beginning all features are regarded as unimportant features. In the training procedure, the gate parameters are updated along with system parameters using full batch GD algorithm. The gate values over $1,000$ iterations are shown in Fig. \ref{wine_gate_plot}. For Fig. \ref{Wine_FS_old_gate} and \ref{Wine_FS_new_gate}, the learning coefficients are set to $0.01$. As expected from Fig. \ref{gatefun_derivative}, the gates of features based on \eqref{new_gate_function} have opened faster and are scattered over a wide range (Fig. \ref{Wine_FS_new_gate}) compared to the other gate function (Fig. \ref{Wine_FS_old_gate}). Consequently, the proposed 
	gate function can distinguish between the useful and poor features faster than the other gate function. However, Fig. \ref{Wine_FS_old_gate} may give a false impression that $M(\lambda)=e^{-\lambda^2}$ cannot do its intended job. This is not true because either with a higher learning coefficient or with more iterations, $M(\lambda)=e^{-\lambda^2}$ can also do its job. To demonstrate this, in Fig. \ref{Wine_FS_old_gate_lr_005} we depict the gate openings with the learning coefficient $0.05$. However, since \eqref{new_gate_function} has a significantly higher magnitude of its derivative with respect to $\lambda$ near the origin, for a given learning coefficient, it will exhibit a faster gate opening, which is an advantage.

	\subsection{AdaTSK Classifier}\label{AdaTSK_classifier}
	\begin{table}[t]
		\setlength\tabcolsep{3pt}
		\caption{Classification accuracy (\%) of TSK fuzzy system with product T-norm, softmin and Ada-softmin}
		\begin{center}
			\begin{tabular}{c c c c}
				\hline
				Dataset (\#Features) & Product & Softmin \scriptsize{($q=-12$)}\cite{2004PalA} & Ada-softmin (ours)\\
				\hline
				Iris (4) & 96.9 & 95.3 & 95.5\\
				Wine (13) & 98.8 & 98.7 & 98.7\\
				Sonar (60) & 75.3 & - & 75.1\\
				ORL (1024) & - & 92.6 & 93.0\\
				Colon (2000) & - & - & 60.0\\
				SRBCT (2308) & - & - & 87.5\\
				ARP (2400) & - & - & 97.5\\
				PIE (2420) & - & - & 98.0\\
				Leukemia (7129) & - & - & 80.0\\
				CNS (7129) & - & - & 60.6\\ \hline
			\end{tabular}
		\end{center}
		\label{adapsoftminresults}
	\end{table}
	
	In order to demonstrate the limitations of product and softmin operators and the advantage of the proposed Ada-softmin, we implement the CoCo-FRB based TSK fuzzy systems for classification on $10$ datasets ($3$ low-dimensional datasets and $7$ high-dimensional datasets). Since our objective is just to demonstrate the effectiveness of AdaTSK framework, no feature selection is done here.

	The average classification accuracy of $10$ repeated experiments is shown in Table \ref{adapsoftminresults}, where '-' indicates that the TSK fuzzy system with product T-norm or softmin fails to get results because  of ``numeric underflow'' during  the training phase. For the product T-norm, the TSK fuzzy classifier  can obtain  excellent results on low-dimensional datasets, but it can not work at all on high-dimensional datasets. On the other hand, the TSK fuzzy classifier with softmin, using parameter $q=-12$ \cite{2004PalA}, can obtain very competitive performance on Iris and Wine datasets but fails to deal with the Sonar dataset. It is interesting to observe that for the ORL, the TSK classifier works quite well with softmin, but product fails for this dataset. One may wonder why softmin did not work for Sonar but performed well on ORL! The answer may lie with the structure of the data and the domain of each feature. Note that, for the results reported in Table \ref{adapsoftminresults} we did not do any feature selection or rule extraction and we have used $S=3$ rules with the CoCo-FRB based AdaTSK model.

	Since the parameter of Ada-softmin, $\hat{q}$, can be adaptively acquired by \eqref{qhat} and the lower bound of $\hat{q}$ is set to $1,000$, the ``numeric underflow'' and ``fake minimum'' mentioned in Subsection \ref{ada_softmin} are avoided. As expected, the proposed AdaTSK can successfully classify all $10$ datasets including seven high-dimensional datasets.  
	
	\begin{table}[t]
		\caption{Results of the low-dimensional datasets}
		\begin{center}
			\begin{tabular}{c c c}
				\hline
				& Previous Work \cite{2018PalFeature} 	& FSRE-AdaTSK\\
				Dataset (\#Features) & Acc (\#F, \#R) 		& Acc (\#F, \#R)\\
				\hline
				Iris (4) 			& 96.0 (1.0, 5.4) 	& \textbf{96.5} (2.1, 6.1)\\
				Appendicities (7) 	& 84.6 (3.8, 3.4)  	& \textbf{86.3} (3.7, 3.1)\\
				Pima (8) 			& 75.4 (4.5, 10.8) 	& \textbf{75.7} (3.0, 3.2)\\
				Yeast (8) 			& \textbf{59.5} (7.5, 16.4) 	& 58.1 (4.5, 10.0)\\
				Glass (9) 			& 65.2 (7.1, 9.6) 	& \textbf{65.5} (4.1, 6.0)\\
				Page-blocks (10) 	& \textbf{94.2} (5.5, 6.0) 	& 93.4 (2.2, 5.6)\\
				Wine (13) 			& 96.7 (4.2, 7.6) 	& \textbf{97.3} (6.3, 6.3)\\
				Heart (13) 			& 80.4 (7.3, 12.7) 	& \textbf{82.5} (7.4, 2.9)\\
				Wdbc (30) 			& 94.6 (2.98, 7.7) 	& \textbf{95.4} (6.2, 4.9)\\
				Texture (40) 		& 94.3 (14.8, 30.3)	& \textbf{97.2} (23.7, 11.0)\\
				Spectfheart (44) 	& \textbf{79.7} (3.3, 4.0)	& 79.2 (3.4, 5.1)\\
				Sonar (60) 			& \textbf{76.7} (8.3, 7.4) 	& 73.4 (10.5, 2.1)\\
				\hline
			\end{tabular}
		\end{center}
		\label{result_low_dimension}
	\end{table}
	
	\begin{table}[!t]
		\caption{Results of the high-dimensional datasets}
		\begin{center}
			\begin{tabular}{c c c}
				\hline
				& FSRE-AdaTSK (LSE) 	& FSRE-AdaTSK (GD)\\
				Dataset (\#Features) & Acc (\#F, \#R) & Acc (\#F, \#R)\\
				\hline
				ORL (1024) 		& \textbf{86.7} (63.7, 40.0) & 83.5 (63.7, 40.0)\\
				Colon (2000) 	& 77.0 (9.0, 6.2) 	& \textbf{81.6} (9.0, 6.2)\\
				SRBCT (2308) 	& 95.1 (12.3, 4.2) 	& \textbf{96.7} (12.3, 4.2)\\ 
				ARP (2400) 		& \textbf{79.1} (45.0, 76.4)	& 77.7 (45.0, 76.4)\\
				PIE (2420) 		& 95.1 (37.9, 10.1)	& \textbf{96.2} (37.9, 10.1)\\
				Leukemia (7129) & 92.3 (11.5, 4.2) 	& \textbf{93.4} (11.5, 4.2)\\
				CNS (7129) 		& 63.0 (51.9, 15.1) & \textbf{71.0} (51.9, 15.1)\\
				\hline
			\end{tabular}
		\end{center}
		\label{result_high_dimension}
	\end{table}
	
	\begin{figure*}[t]
		\centering
		\subfigure[Feature selection phase]{\includegraphics[scale=0.55]{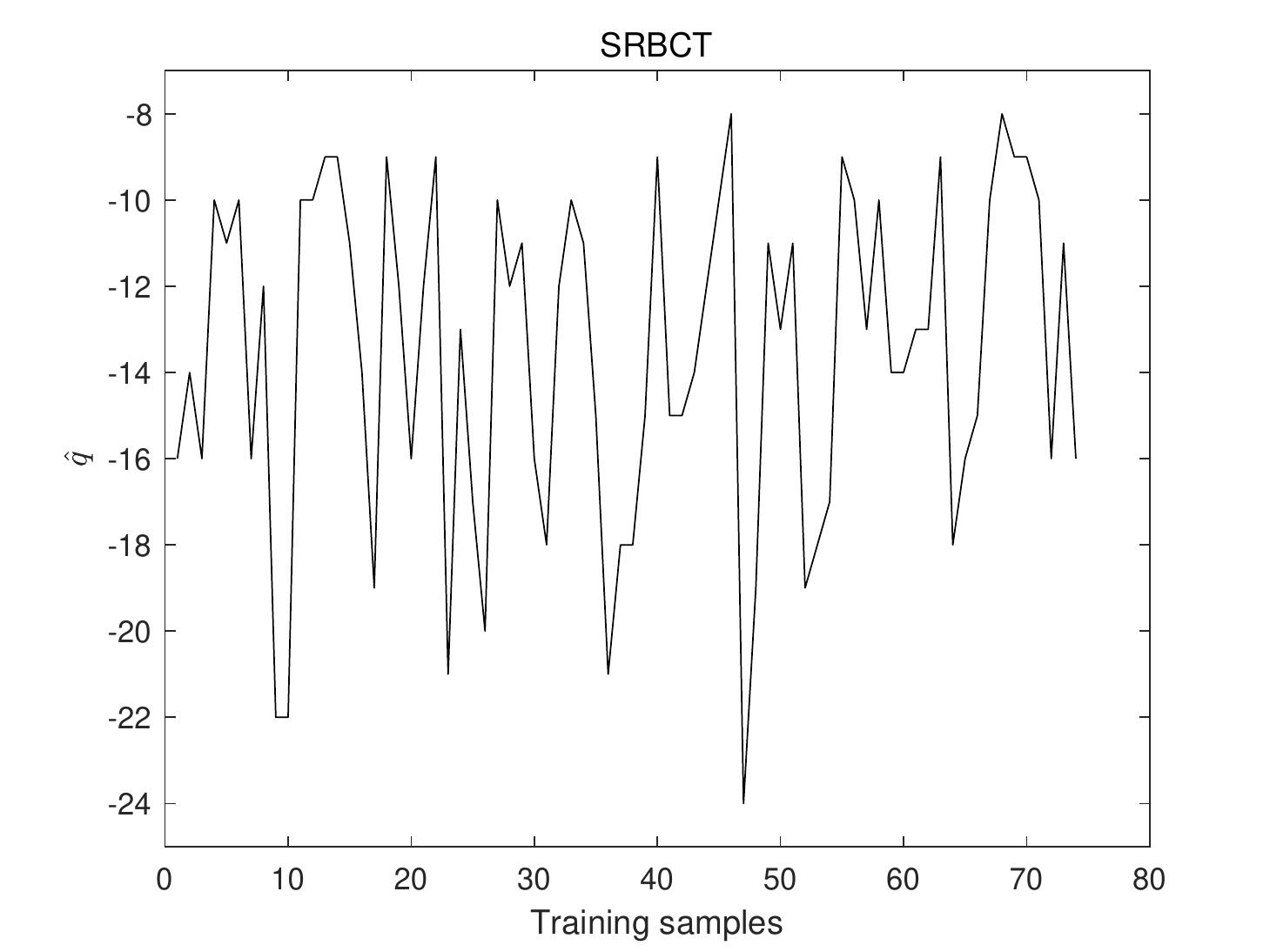}\label{SRBCT_q_FS}}
		\subfigure[Rule extraction phase]{\includegraphics[scale=0.55]{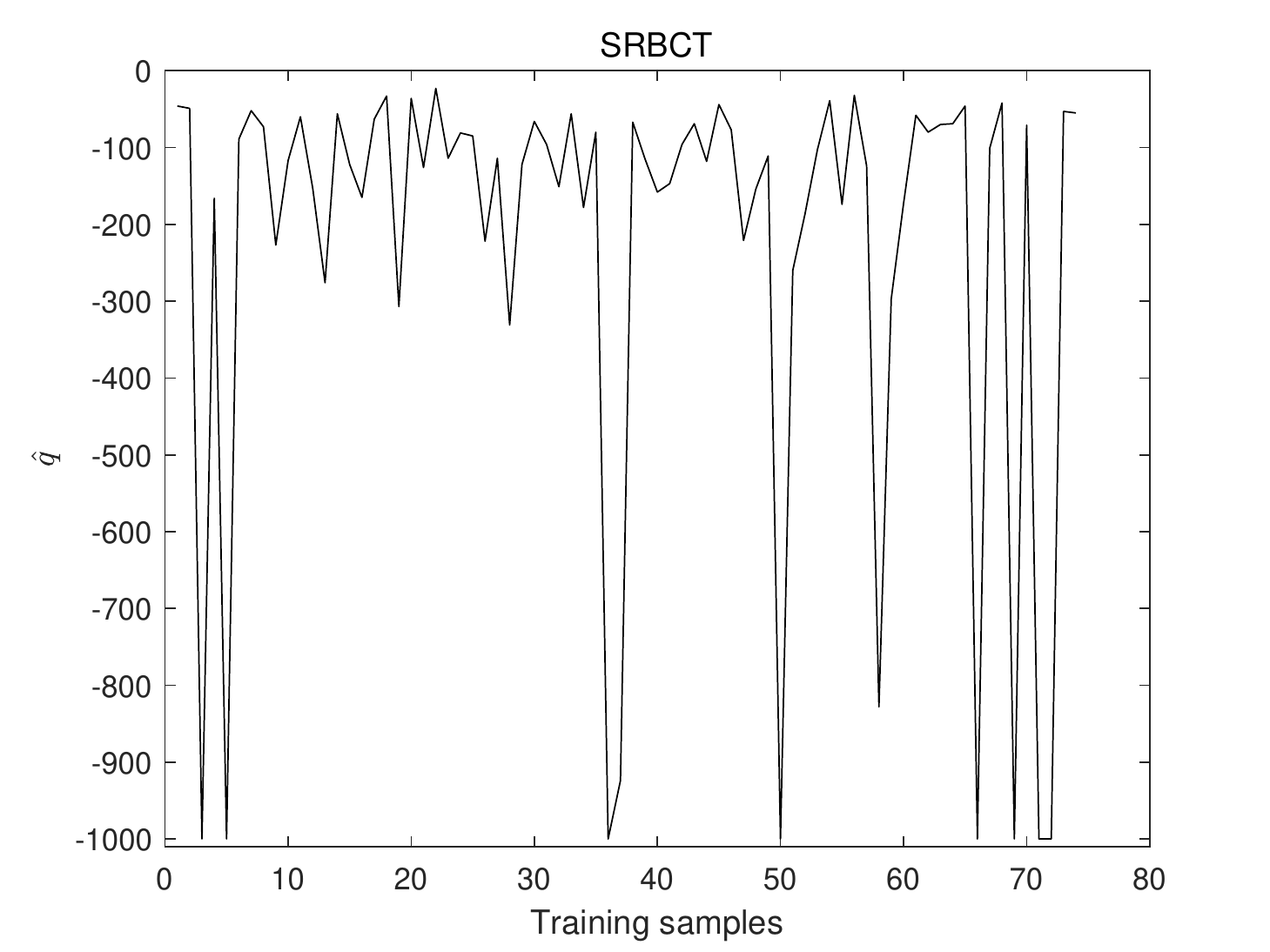}\label{SRBCT_q_RE}}
		\vfil
		\subfigure[Fine tuning phase]
		{\includegraphics[scale=0.55]{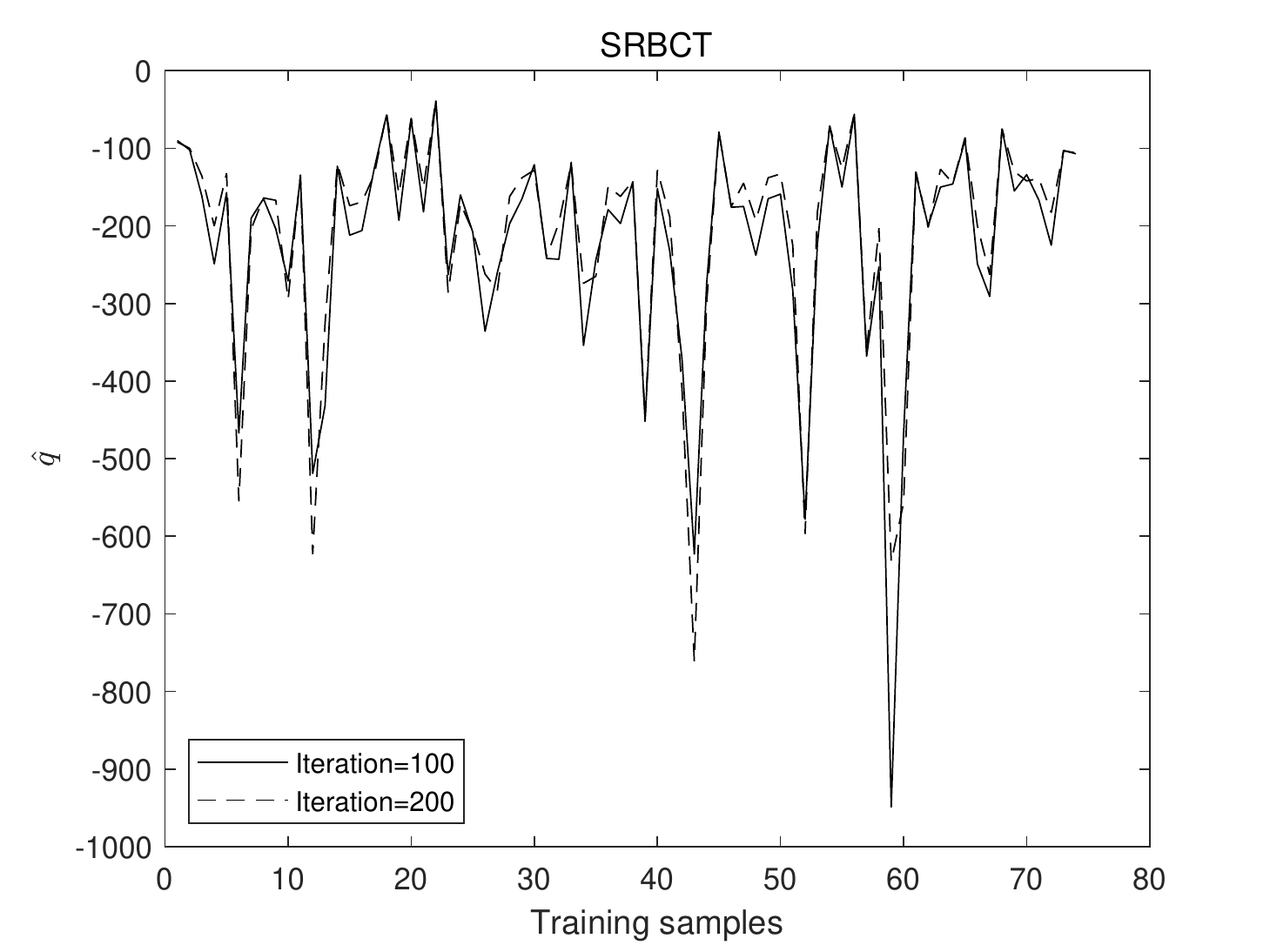}\label{SRBCT_q_ini_FT}}
		\subfigure[The changes of $\hat{q}$ with iterations in Fine tuning phase]
		{\includegraphics[scale=0.55]{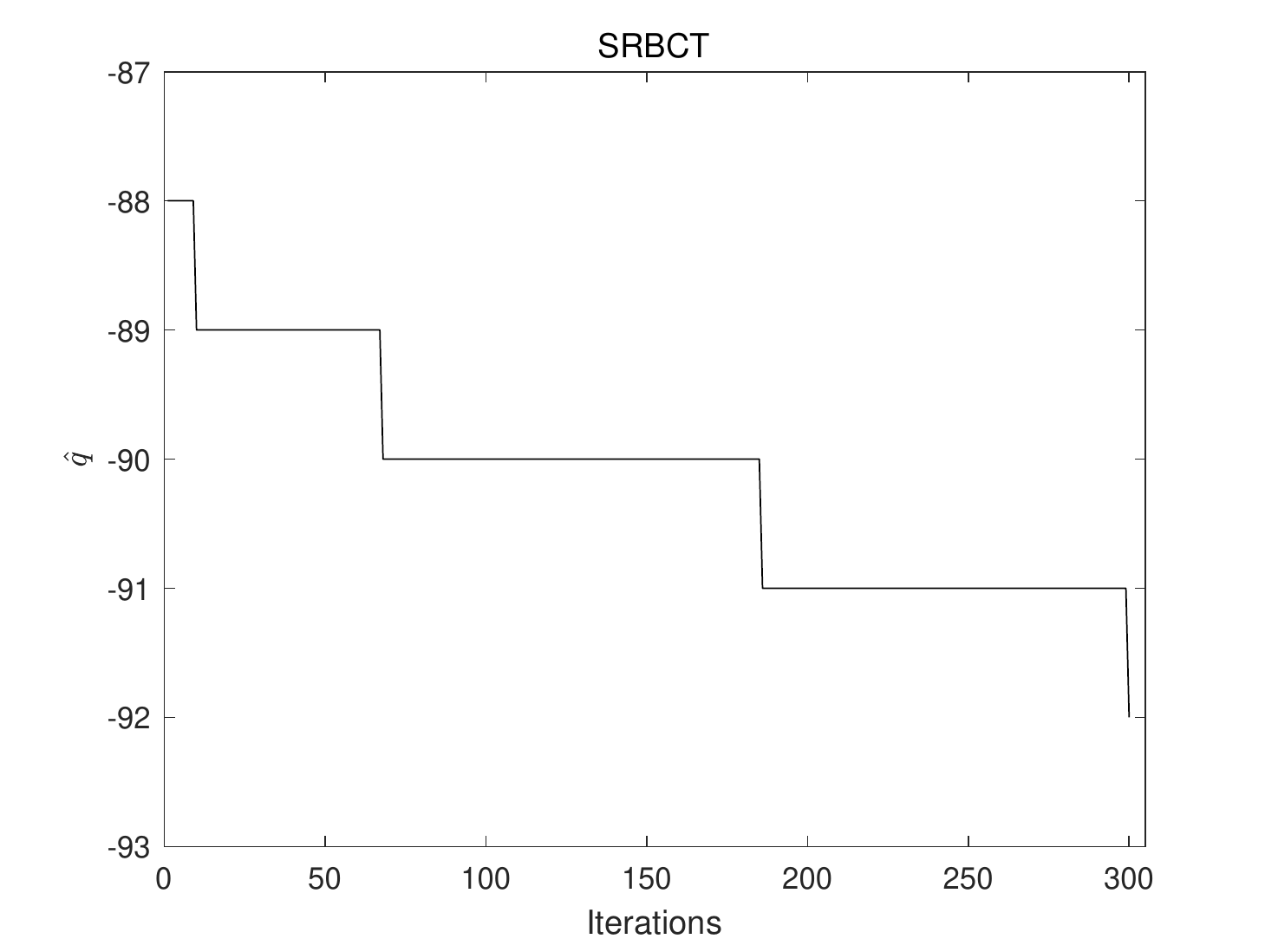}\label{SRBCT_q_FT}}
		\caption{The $\hat{q}$ of Ada-softmin used on SRBCT dataset}
		\label{SRBCT_q}
	\end{figure*}

	\subsection{The Classification Performance of FSRE-AdaTSK}\label{FSRE_AdaTSK_results}
	To demonstrate the efficient performance of FSRE-AdaTSK, we test it on $19$ datasets with dimension varying from $4$ to $7129$. 
	As explained earlier, there are three phases of the entire learning process. In the first phase, the feature selection phase, we use $10$ fuzzy sets defined on every feature. The use of a large number of fuzzy sets is motivated by the fact that a high resolution fuzzy partition may help the feature selection process. We admit that for data sets like Iris, we do not need $10$ fuzzy sets, but just to have a uniform policy for all data sets, we follow this.  Moreover, in this phase we use the CoCo-FRB, i.e., use only $R=S=10$ rules. Thus, use of $10$ rules will help model classes with complex structure. This will also indirectly reveal the robustness of the feature selection process. For the rule extraction phase, we define $5$ linguistic values or fuzzy sets on each feature with a hope that since we already have identified useful features we may not need high resolution fuzzy partition. On the other hand, since in this phase we want to extract the final rules, we begin the process with more rules. In particular, in this phase, we use En-FRB, i.e, we begin with $R=(2D'+1)S=(2D'+1)*5$ rules, where $D'$ is the number of selected features in the previous phase.
	
	For the feature selection phase, to guarantee that all features are regarded as unimportant features at the very beginning of training, gate parameter of every feature is initialized to $0.01$, i.e., every gate value (\eqref{new_gate_function} is used here) is initialized to $0.0165$. Similarly, at the beginning of the rule extraction phase, every rule gate parameter is set to $0.01$. The initialization method of system parameters is identical to that described earlier in the beginning of Section \ref{experiments_results}. Since the feature selection and rule extraction are independent of each other, the system parameters and gate parameters for rule extraction must be reinitialized after the feature selection procedure. While in the fine tuning phase, all of the parameters do not need to be reinitialized. In our subsequent discussion we use $R'$ as the number of rules finally extracted from the second phase, which will be used in the final refinement phase.

	As explained earlier, the antecedent parameters are not updated in the first two phases when handling with high-dimensional datasets. While in the fine tuning phase, based on the reduced feature space and the simplified fuzzy rule base, all of the system parameters including centers and consequent parameters are tuned. Ten-fold cross-validation mechanism is also used here and it is repeated $10$ times as well. The average classification accuracy (Acc), average number of selected features (\#F) and average number of extracted rules (\#R) are reported in Table \ref{result_low_dimension} and Table \ref{result_high_dimension} for the low and high-dimensional datasets, respectively. 

	For low-dimensional datasets, shown in Table \ref{result_low_dimension}, we compare the results of FSRE-AdaTSK and those of previous work \cite{2018PalFeature}. In \cite{2018PalFeature}, the authors used gate function $M(\lambda)=e^{-\lambda^2}$, based on a fuzzy rule based framework, to select features with controlled redundancy for solving classification problems in which the dimensions of used datasets are all less than $1,000$. As seen from Table \ref{result_low_dimension}, the proposed scheme, FSRE-AdaTSK, uses fewer rules for most of the datasets and gets comparative results. Especially on some datasets, such as Appendicities and Glass, FSRE-AdaTSK obtains higher classification accuracy using fewer features and rules.
	
	Since we could not find any existing fuzzy method to deal with classification problems with dimension more than 1000, would could not compare the performance of FSRE-AdaTSK with any other fuzzy methods. However, in Table \ref{result_high_dimension} we present results using two kinds of parameter estimation methods for FSRE-AdaTSK scheme. Different from aforementioned GD based method, in the fine tuning phase, the consequent parameters are obtained using  the least square error (LSE) estimation method while the antecedent parameters are kept fixed. The equation for the LSE estimate is not included here as this is not the primary research focus here. In \cite{1006431}, one can find the elaborate procedure for the LSE estimation of the consequent parameters. From Table \ref{result_high_dimension}, it can be concluded that FSRE-AdaTSK scheme is capable of solving high-dimensional problems and acquiring very satisfactory results regardless of whether we use the LSE estimation method or full batch GD algorithm in the fine tuning phase.
	
	To illustrate that the proposed Ada-softmin can adaptively acquire $\hat{q}$ according to the current membership values, in Fig. \ref{SRBCT_q} we show the $\hat{q}$ used in different phases while dealing with the dataset, SRBCT. For dealing with high-dimensional datasets, the centers are kept fixed in the first two phases: feature selection and rule extraction. The parameters of Ada-softmin are found to adapt to different $\hat{q}$s for different instances before the training and maintained these values during the training. Note that, for every rule, there will be a set computed values of  $\hat{q}$s, one for each instance. For a typical rule, the computed $\hat{q}$s corresponding to all the training instances of SRBCT in the feature selection and rule extraction phases are shown in Fig. \ref{SRBCT_q_FS} and Fig. \ref{SRBCT_q_RE}, respectively. On the other hand, in the fine tuning phase, the centers and the consequent parameters are trained together and consequently, the $\hat{q}$s are changed along with iterations. The $\hat{q}$s calculated for a typical rule for different instances just before the $100_{th}$ and the $200_{th}$ iterations are shown in Fig. \ref{SRBCT_q_ini_FT}. This figure depicts how the $\hat{q}$ changes with iterations for different instances. Moreover, for a given rule and for one typical instance, the changes of $\hat{q}$ along with iterations are plotted in Fig. \ref{SRBCT_q_FT}.

	\section{Conclusion}\label{conclusion}
	In this paper, we propose a comprehensive fuzzy rule-based scheme called as FSRE-AdaTSK to solve high-dimensional classification problems. In order to address the problems associated with the computation of firing strength for high-dimensional data, we propose an adaptive softmin, called,  Ada-softmin.  The TSK model using Ada-Softmin is named  here   AdaTSK.  The AdaTSK fuzzy system is capable of dealing with high-dimensional datasets. In addition, to facilitate feature selection and rule extraction using the  AdaTSK framework, we propose a novel gate function, which is embedded into the system. This new gate function eliminates a limitation of the commonly used gate functions for feature selection. The feature selection and rule extraction are done  in two different phases and for each phase we use a separate set of gate functions. The existing embedded methods for fuzzy rule based feature selection use the gate function in the rule antecedents for the MA model and both in the antecedents and consequents for the TSK model. But the proposed system uses the gate function only with consequents both for feature selection and rule extraction.   In the rule extraction/selection phase, we design a new type of fuzzy rule base, En-FRB, which begins with more rules than CoCo-FRB and avoids the exponential growth of number of rules with dimensions for FuCo-FRB. 

	We demonstrate the effectiveness of the proposed system on  $19$ datasets of which seven datasets have dimension between $1000$ and $7129$. To the best of our knowledge, this is the first time fuzzy rule based systems have been designed involving data of dimension more than $7000$, where the feature selection is also done using a fuzzy rule based framework. We note that both of the feature selection and rule extraction processes are based on the first-order TSK fuzzy system, so the number of system parameters that need to be optimized is quite large. In order to reduce the training burden, the zero-order TSK model will be considered in our future work. Use of the proposed AdaTSK framework for regression/prediction problems is straightforward. We plan to check the effectiveness of AdaTSK for function approximation/regression type problems.  Besides,  designing more efficient strategies for identifying fuzzy systems involving  high-dimensional problems  deserves further investigation.


\begin{thebibliography}{10}
		
		\bibitem{1975MamdaniAn}
		E.~H. Mamdani and S.~Assilian, ``An experiment in linguistic synthesis with a
		fuzzy logic controller,'' {\em International Journal of Man-Machine Studies},
		vol.~7, no.~1, pp.~1--13, 1975.
		
		\bibitem{1985TakagiFuzzy}
		T.~{Takagi} and M.~{Sugeno}, ``Fuzzy identification of systems and its
		applications to modeling and control,'' {\em IEEE Transactions on Systems,
			Man, and Cybernetics}, vol.~SMC-15, pp.~116--132, Jan 1985.
		
		\bibitem{1986SugenoFuzzy}
		M.~{Sugeno} and G.~T. {Kang}, ``Fuzzy modelling and control of multilayer
		incinerator,'' {\em Fuzzy Sets \& Systems}, vol.~18, no.~3, pp.~329--345,
		1986.
		
		\bibitem{2001PalIntegrated}
		D.~{Chakraborty} and N.~R. {Pal}, ``Integrated feature analysis and fuzzy
		rule-based system identification in a neuro-fuzzy paradigm,'' {\em IEEE
			Transactions on Systems, Man, and Cybernetics, Part B (Cybernetics)},
		vol.~31, pp.~391--400, June 2001.
		
		\bibitem{2004PalA}
		D.~{Chakraborty} and N.~R. {Pal}, ``A neuro-fuzzy scheme for simultaneous
		feature selection and fuzzy rule-based classification,'' {\em IEEE
			Transactions on Neural Networks}, vol.~15, pp.~110--123, Jan 2004.
		
		\bibitem{2012PalAnIntegrated}
		Y.~{Chen}, N.~R. {Pal}, and I.~{Chung}, ``An integrated mechanism for feature
		selection and fuzzy rule extraction for classification,'' {\em IEEE
			Transactions on Fuzzy Systems}, vol.~20, pp.~683--698, Aug 2012.
		
		\bibitem{2018PalFeature}
		I.~{Chung}, Y.~{Chen}, and N.~R. {Pal}, ``Feature selection with controlled
		redundancy in a fuzzy rule based framework,'' {\em IEEE Transactions on Fuzzy
			Systems}, vol.~26, pp.~734--748, April 2018.
		
		\bibitem{2019GaoConjugate}
		T.~Gao, Z.~Zhang, Q.~Chang, X.~Xie, and J.~Wang, ``Conjugate gradient-based
		takagi-sugeno fuzzy neural network parameter identification and its
		convergence analysis,'' {\em Neurocomputing}, vol.~364, pp.~168--181, 2019.
		
		\bibitem{2020WuOptimize}
		D.~{Wu}, Y.~{Yuan}, J.~{Huang}, and Y.~{Tan}, ``{Optimize TSK Fuzzy Systems for
			Regression Problems: Minibatch Gradient Descent With Regularization,
			DropRule, and AdaBound (MBGD-RDA)},'' {\em IEEE Transactions on Fuzzy
			Systems}, vol.~28, pp.~1003--1015, May 2020.
		
		\bibitem{2020CuiOptimize}
		Y.~Cui, D.~Wu, and J.~Huang, ``{Optimize TSK Fuzzy Systems for Classification
			Problems: Minibatch Gradient Descent With Uniform Regularization and Batch
			Normalization},'' {\em IEEE Transactions on Fuzzy Systems}, vol.~28, no.~12,
		pp.~3065--3075, 2020.
		
		\bibitem{1992HorikawaOn}
		S.~Horikawa, T.~Furuhashi, and Y.~Uchikawa, ``On fuzzy modeling using fuzzy
		neural networks with the back-propagation algorithm,'' {\em IEEE Transactions
			on Neural Networks}, vol.~3, no.~5, pp.~801--806, 1992.
		
		\bibitem{1986RumelhartLearning}
		D.~E. Rumelhart, G.~E. Hinton, and R.~J. Williams, ``Learning representations
		by back-propagating errors,'' {\em Nature}, vol.~323, no.~6088, pp.~533--536,
		1986.
		
		\bibitem{1995KlirFuzzy}
		G.~J. Klir and B.~Yuan, {\em Fuzzy sets and fuzzy logic - theory and
			applications}.
		\newblock Prentice Hall, New Jersey, 1995.
		
		\bibitem{1989MizumotoPictorial}
		M.~Mizumoto, ``Pictorial representations of fuzzy connectives, part {I}: Cases
		of t-norms, t-conorms and averaging operators,'' {\em Fuzzy Sets \& Systems},
		vol.~31, no.~2, pp.~217--242, 1989.
		
		\bibitem{2021GuFast}
		G.~Suhang, C.~M. Vong, P.~K. Wong, and S.~Wang, ``Fast training of adversarial
		deep fuzzy classifier by downsizing fuzzy rules with gradient guided
		learning,'' {\em IEEE Transactions on Fuzzy Systems}, pp.~1--1, 2021.
		
		\bibitem{2008PalSimultaneous}
		N.~R. {Pal} and S.~{Saha}, ``Simultaneous structure identification and fuzzy
		rule generation for takagi-sugeno models,'' {\em IEEE Transactions on
			Systems, Man, and Cybernetics, Part B (Cybernetics)}, vol.~38,
		pp.~1626--1638, Dec 2008.
		
		\bibitem{2021WangFeature}
		J.~Wang, H.~Zhang, J.~Wang, Y.~Pu, and N.~R. Pal, ``Feature selection using a
		neural network with group lasso regularization and controlled redundancy,''
		{\em IEEE Transactions on Neural Networks and Learning Systems}, vol.~32,
		no.~3, pp.~1110--1123, 2021.
		
		\bibitem{2017SolorioANew}
		S.~{Solorio-Fernández}, J.~F. {Martínez-Trinidad}, and J.~A.
		{Carrasco-Ochoa}, ``A new unsupervised spectral feature selection method for
		mixed data: A filter approach,'' {\em Pattern Recognition}, vol.~72,
		pp.~314--326, 2017.
		
		\bibitem{1997KohaviWrappers}
		R.~{Kohavi} and G.~H. {John}, ``Wrappers for feature subset selection,'' {\em
			Artificial Intelligence}, vol.~97, no.~1, pp.~273--324, 1997.
		
		\bibitem{2009MaldonadoA}
		S.~Maldonado and R.~Weber, ``A wrapper method for feature selection using
		support vector machines,'' {\em Information Sciences}, vol.~179, no.~13,
		pp.~2208--2217, 2009.
		
		\bibitem{2010KabirA}
		M.~M. {Kabir}, M.~M. {Islam}, and K.~{Murase}, ``A new wrapper feature
		selection approach using neural network,'' {\em Neurocomputing}, vol.~73,
		no.~16, pp.~3273--3283, 2010.
		
		\bibitem{2011HsuHybrid}
		H.-H. {Hsu}, C.-W. {Hsieh}, and M.-D. {Lu}, ``Hybrid feature selection by
		combining filters and wrappers,'' {\em Expert Systems With Applications},
		vol.~38, no.~7, pp.~8144--8150, 2011.
		
		\bibitem{2019ChenAn}
		J.~Chen, T.~Li, Y.~Zou, G.~Wang, H.~Ye, and F.~Lv, ``An ensemble feature
		selection method for short-term electrical load forecasting,'' in {\em 2019
			IEEE 3rd Conference on Energy Internet and Energy System Integration (EI2)},
		pp.~1429--1432, 2019.
		
		\bibitem{2015HuHybrid}
		Z.~{Hu}, Y.~{Bao}, T.~{Xiong}, and R.~{Chiong}, ``Hybrid filter-wrapper feature
		selection for short-term load forecasting,'' {\em Engineering Applications of
			Artificial Intelligence}, vol.~40, pp.~17--27, 2015.
		
		\bibitem{2018LiuA}
		X.-Y. Liu, Y.~Liang, S.~Wang, Z.-Y. Yang, and H.-S. Ye, ``A hybrid genetic
		algorithm with wrapper-embedded approaches for feature selection,'' {\em IEEE
			Access}, vol.~6, pp.~22863--22874, 2018.
		
		\bibitem{2008PalSelecting}
		D.~{Chakraborty} and N.~R. {Pal}, ``Selecting useful groups of features in a
		connectionist framework,'' {\em IEEE Transactions on Neural Networks},
		vol.~19, pp.~381--396, March 2008.
		
		\bibitem{2015PalFeature}
		R.~{Chakraborty} and N.~R. {Pal}, ``Feature selection using a neural framework
		with controlled redundancy,'' {\em IEEE Transactions on Neural Networks and
			Learning Systems}, vol.~26, pp.~35--50, Jan 2015.
		
		\bibitem{2020ZhangFeature}
		H.~Zhang, J.~Wang, Z.~Sun, J.~M. Zurada, and N.~R. Pal, ``Feature selection for
		neural networks using group lasso regularization,'' {\em IEEE Transactions on
			Knowledge and Data Engineering}, vol.~32, no.~4, pp.~659--673, 2020.
		
		\bibitem{1995AbeA}
		S.~Abe and M.-S. Lan, ``A method for fuzzy rules extraction directly from
		numerical data and its application to pattern classification,'' {\em IEEE
			Transactions on Fuzzy Systems}, vol.~3, no.~1, pp.~18--28, 1995.
		
		\bibitem{1992IshibuchiDistributed}
		H.~Ishibuchi, K.~Nozaki, and H.~Tanaka, ``Distributed representation of fuzzy
		rules and its application to pattern classification,'' {\em Fuzzy sets and
			systems}, vol.~52, no.~1, pp.~21--32, 1992.
		
		\bibitem{1995IshibuchiSelecting}
		H.~{Ishibuchi}, K.~{Nozaki}, N.~{Yamamoto}, and H.~{Tanaka}, ``Selecting fuzzy
		if-then rules for classification problems using genetic algorithms,'' {\em
			IEEE Transactions on Fuzzy Systems}, vol.~3, pp.~260--270, Aug 1995.
		
		\bibitem{2004IshibuchiFuzzy}
		H.~Ishibuchi and T.~Yamamoto, ``Fuzzy rule selection by multi-objective genetic
		local search algorithms and rule evaluation measures in data mining,'' {\em
			Fuzzy Sets \& Systems}, vol.~141, no.~1, pp.~59--88, 2004.
		
		\bibitem{2006IshibuchiFuzzy}
		H.~{Ishibuchi}, Y.~{Nojima}, and I.~{Kuwajima}, ``Fuzzy data mining by
		heuristic rule extraction and multiobjective genetic rule selection,'' in
		{\em 2006 IEEE International Conference on Fuzzy Systems}, pp.~1633--1640,
		July 2006.
		
		\bibitem{2014SrivastavaDropout}
		N.~Srivastava, G.~Hinton, A.~Krizhevsky, I.~Sutskever, and R.~Salakhutdinov,
		``Dropout: A simple way to prevent neural networks from overfitting,'' {\em
			Journal of Machine Learning Research}, vol.~15, no.~1, pp.~1929--1958, 2014.
		
		\bibitem{2013WanRegularization}
		L.~Wan, M.~D. Zeiler, S.~Zhang, Y.~Lecun, and R.~Fergus, ``Regularization of
		neural networks using dropconnect,'' in {\em International Conference on
			Machine Learning}, 2013.
		
		\bibitem{8432091}
		S.~{Feng} and C.~L.~P. {Chen}, ``Fuzzy broad learning system: A novel
		neuro-fuzzy model for regression and classification,'' {\em IEEE Transactions
			on Cybernetics}, vol.~50, pp.~414--424, Feb 2020.
		
		\bibitem{2020GuA}
		S.~Gu, F.~Chung, and S.~Wang, ``{A Novel Deep Fuzzy Classifier by Stacking
			Adversarial Interpretable TSK Fuzzy Sub-Classifiers With Smooth Gradient
			Information},'' {\em IEEE Transactions on Fuzzy Systems}, vol.~28, no.~7,
		pp.~1369--1382, 2020.
		
		\bibitem{1992WangBack}
		L.-X. Wang and J.~Mendel, ``Back-propagation fuzzy system as nonlinear dynamic
		system identifiers,'' in {\em [1992 Proceedings] IEEE International
			Conference on Fuzzy Systems}, pp.~1409--1418, 1992.
		
		\bibitem{2012BengioPractical}
		Y.~{Bengio}, ``Practical recommendations for gradient-based training of deep
		architectures,'' {\em arXiv preprint arXiv:1206.5533}, 2012.
		
		\bibitem{1006431}
		N.~R. {Pal}, V.~K. {Eluri}, and G.~K. {Mandal}, ``Fuzzy logic approaches to
		structure preserving dimensionality reduction,'' {\em IEEE Transactions on
			Fuzzy Systems}, vol.~10, pp.~277--286, June 2002.
		
	\end{thebibliography}

	\begin{IEEEbiography}[{\includegraphics[width=1in,height=1.25in,clip,keepaspectratio]{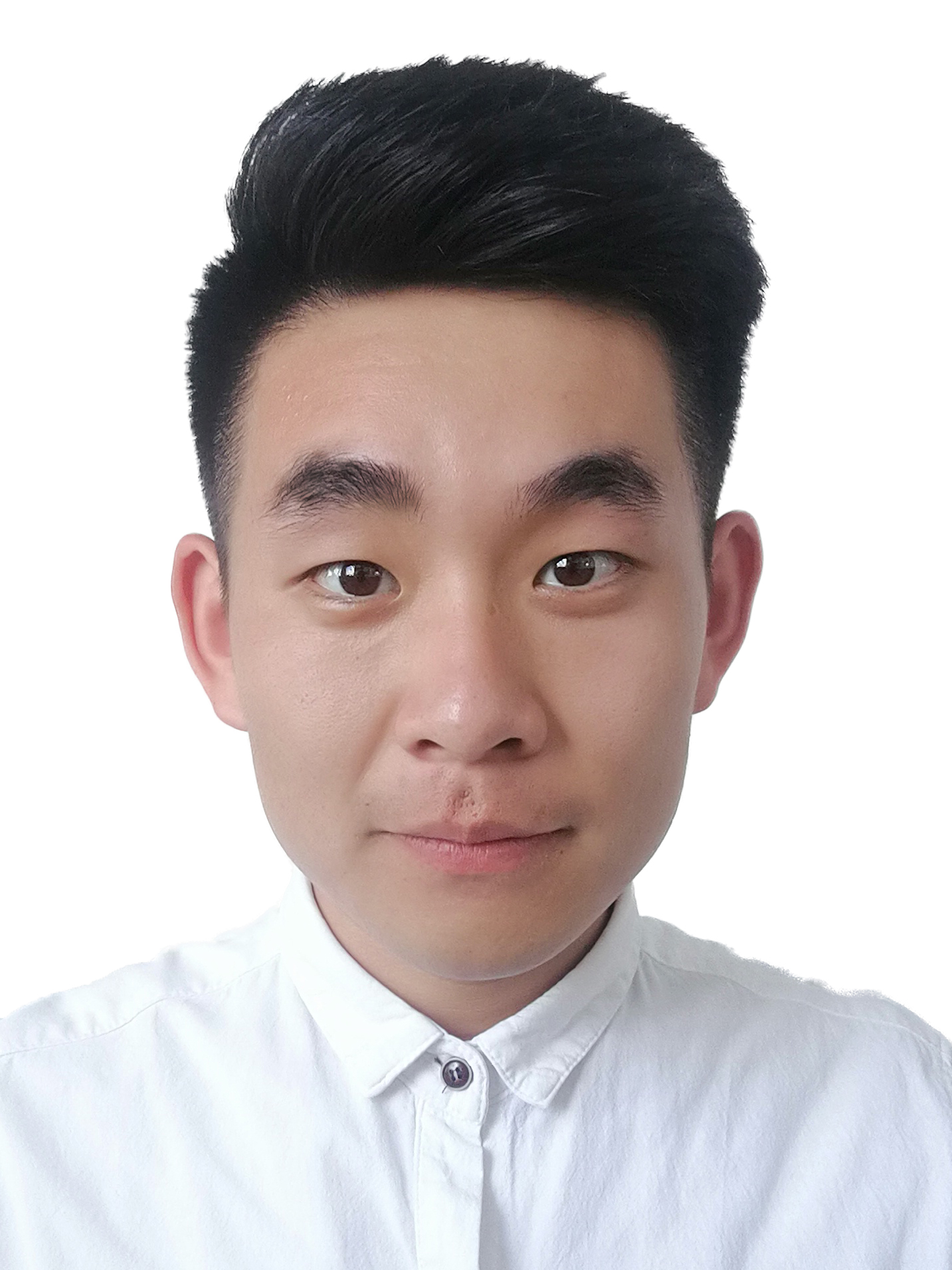}}]{Guangdong Xue}
		received the B.M. degree in logistics management from Hainan University, Haikou, China, in 2018. Currently, he is pursuing the Ph.D. degree of control science and engineering in China University of Petroleum (East China), Qingdao, China. His current research interests include fuzzy systems, pattern recognition and neural networks.
	\end{IEEEbiography}
	
	\begin{IEEEbiography}[{\includegraphics[width=1in,height=1.25in,clip,keepaspectratio]{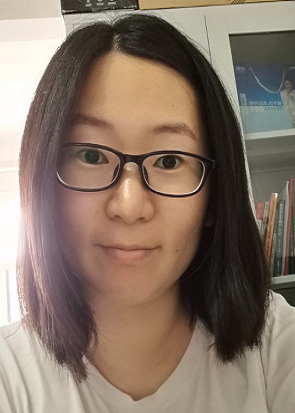}}]{Qin Chang}
		received the B.S. degree in applied mathematics from the China University of Petroleum (East China), Qingdao, China, in 2006, and the Ph.D. degree in probability and statistics from Shandong University, Jinan, China, in 2012.
		
		She joined the China University of Petroleum (East China) in 2013, where she is currently a Lecturer with the Department of Data Science and Statistics, College of Science. Her current research interests include pattern recognition and classification, artificial neural networks, and bioinformatics.
	\end{IEEEbiography}
	
	\begin{IEEEbiography}[{\includegraphics[width=1in,height=1.25in,clip,keepaspectratio]{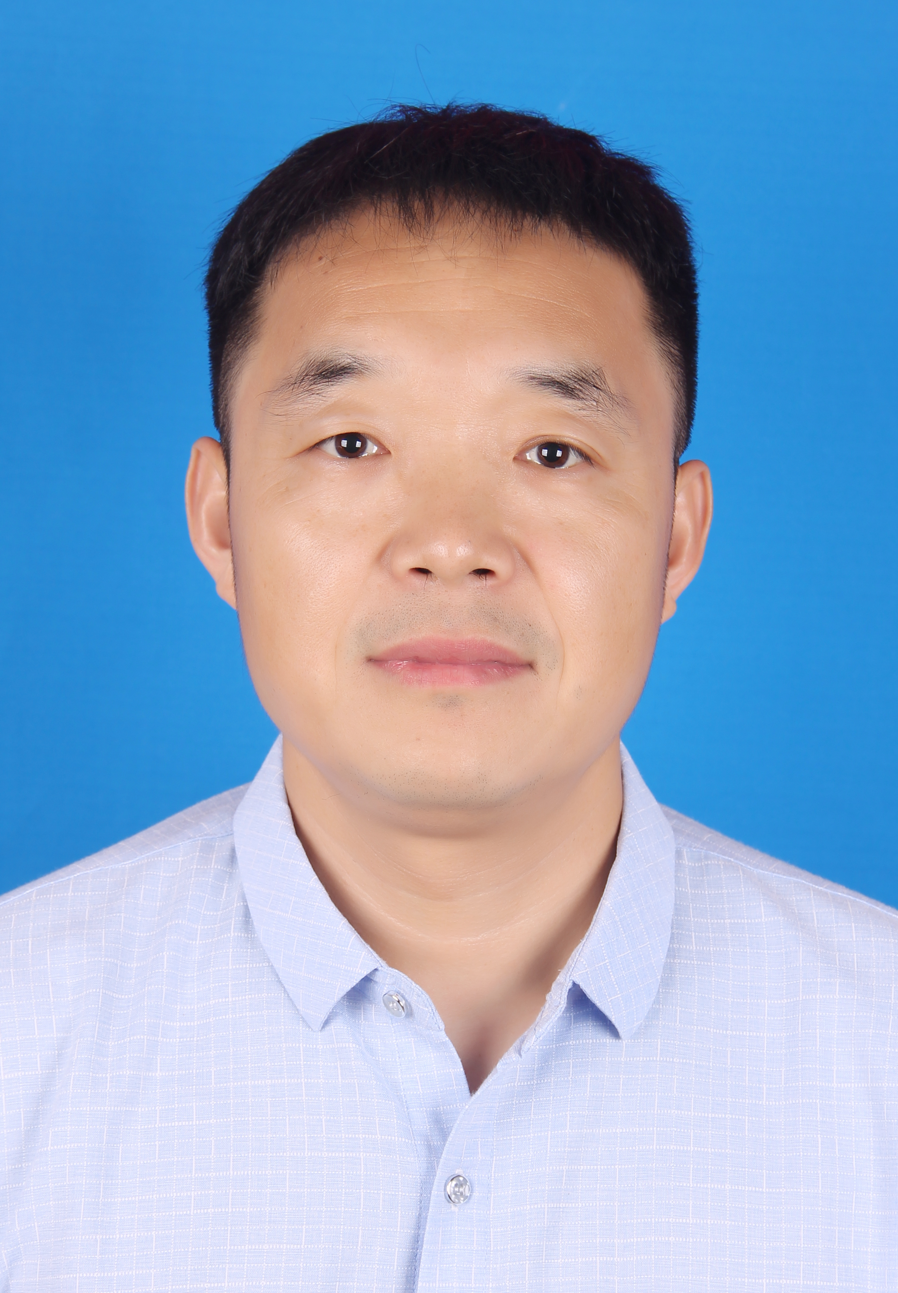}}]{Jian Wang}
		received the B.S. degree in computational mathematics from the China University of Petroleum (East China), Qingdao, China, in 2002, and the M.S. and Ph.D. degrees in computational mathematics from the Dalian University of Technology, Dalian, China, in 2008 and 2012, respectively. 
		
		He is currently a Professor and serves as the Head of the Laboratory for Intelligent Information Processing with the College of Science, China University of Petroleum (East China). His research interests include computational intelligence, machine learning, pattern recognition, deep learning, differential programming, clustering, fuzzy systems, evolutionary computation. He was awarded several grants from the National Science Foundation of China, National Key Research and Development Program of China, Natural Science Foundation of Shandong Province, Fundamental Research Funds for the Central Universities. 
		
		Prof. Wang serves as an Associate Editor for the IEEE Transactions on Neural Networks and Learning Systems, International Journal of Machine Learning and Cybernetics, and Journal of Applied Computer Science Methods. He also serves on the Editorial Board for the Neural Computing \& Applications and Complex \& Intelligent Systems. Prof. Wang served as a Guest Editor for the Neural Computing \& Applications and Computational Intelligence. In addition, He has served as the General Chair, the Program Chair, and the Co-Program Chair of several conferences such as the International Symposium on New Trends in Computational Intelligence, IEEE Symposium Series on Computational Intelligence and International Symposium on Neural Networks.
	\end{IEEEbiography}
	
	\begin{IEEEbiography}[{\includegraphics[width=1in,height=1.25in,clip,keepaspectratio]{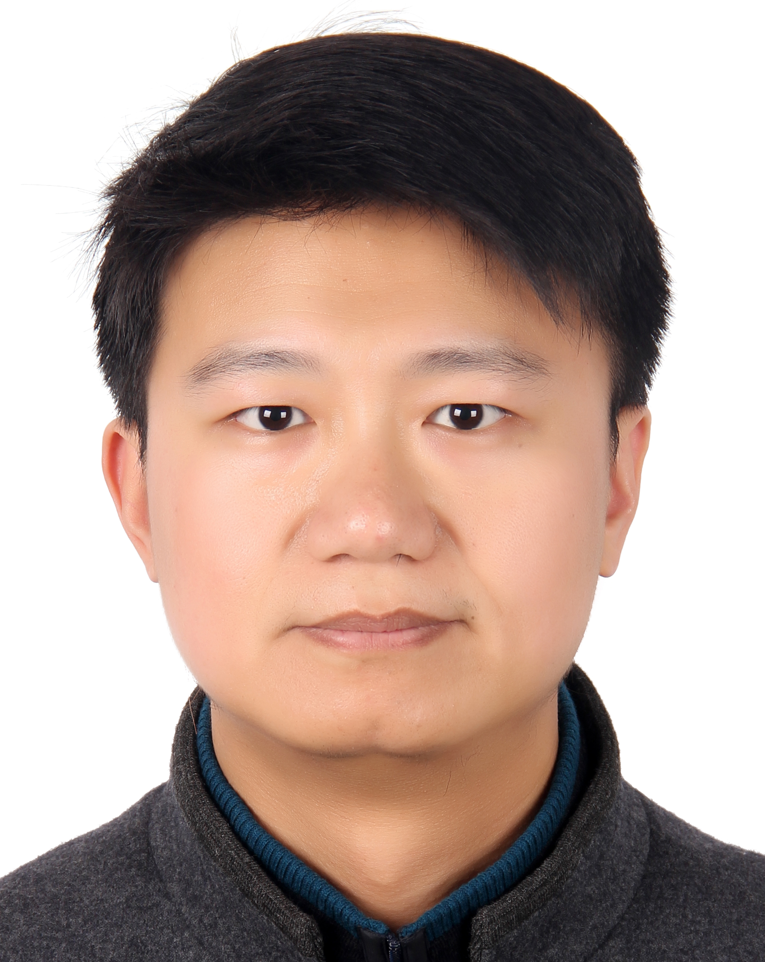}}]{Kai Zhang}
		Ph.D., Professor, Director of Talent Office in China University of Petroleum (East China), SPE member, and InterPore member. From June 2007 to May 2008, he studied in University of Tulsa in the US. After obtaining the Ph.D. degree in China University of Petroleum (East China) in 2008, he began teaching in this university ever since. He teaches courses including Fluid flow in porous media, reservoir engineering etc. His researches focus on reservoir simulation, production optimization, history  matching,  and  development  of nonconventional reservoir etc. As a project leader, he has been in charge of 3 projects supported by the Natural Science Foundation of China (NSFC), 1 project supported by the National Natural Science Foundation of Shandong Province, 1 project supported by the National Ministry of Education, 20 projects supported by SINOPEC, CNOOC, and CNPC. He has already published more than 60 papers.
	\end{IEEEbiography}
	
	\begin{IEEEbiography}[{\includegraphics[width=1in,height=1.25in,clip,keepaspectratio]{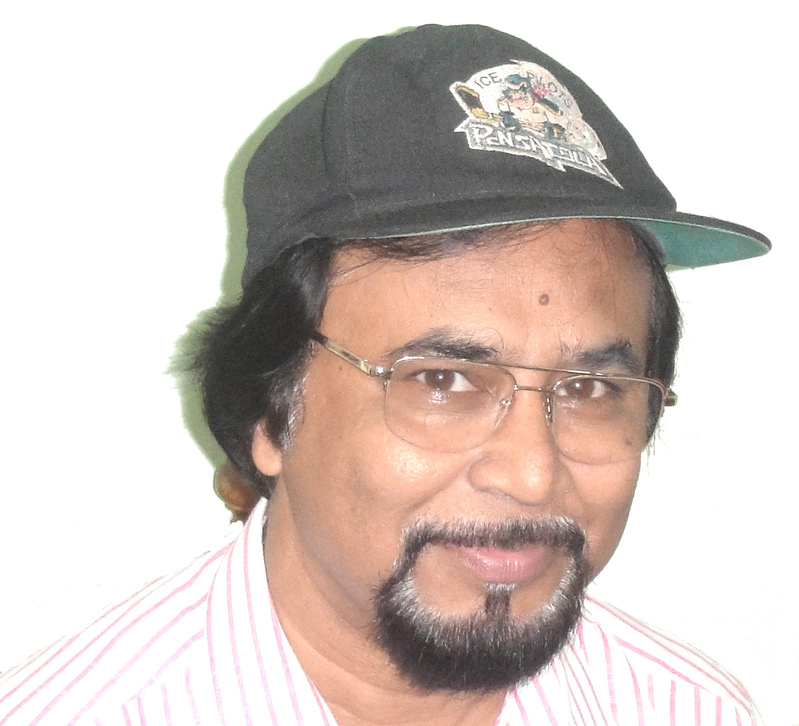}}]{Nikhil R. Pal}
		is a Professor in the Electronics and Communication Sciences Unit  of the Indian Statistical Institute (ISI). At present, he is the Head of the Center for Artificial Intelligence and Machine Learning at ISI.  His current research interest includes brain science, computational intelligence, machine
		learning and data mining.
		
		He was the Editor-in-Chief of the IEEE Transactions on Fuzzy Systems for the period January 2005-December 2010. He has served/been serving on the editorial /advisory board/ steering committee of several journals including the International Journal of Approximate Reasoning, Applied Soft Computing, International Journal of Neural Systems, Fuzzy Sets and Systems, IEEE Transactions on Fuzzy Systems and the IEEE Transactions on Cybernetics.
		
		He is a recipient of the 2015 IEEE Computational Intelligence Society (CIS) Fuzzy Systems Pioneer Award, He has given many plenary/keynote speeches in different premier international conferences in the area of computational intelligence. He has served as the General Chair, Program Chair, and co-Program chair of several conferences. He was a Distinguished Lecturer of the IEEE CIS (2010-2012, 2016-2018.) and was a member of the Administrative Committee of the IEEE CIS (2010-2012). He has served as the Vice-President for Publications of the IEEE CIS (2013-2016) as well as the President of the IEEE CIS (2018-2019).
		
		He is a Fellow of the National Academy of Sciences, India, Indian National Academy of Engineering, Indian National Science Academy, International Fuzzy Systems Association (IFSA), The World Academy of Sciences, and a Fellow of the IEEE, USA.
	\end{IEEEbiography}

\end{document}